\documentclass[10pt]{article}
\usepackage[utf8]{inputenc}
\usepackage[T1]{fontenc}
\usepackage{amsmath}
\usepackage{amsfonts}
\usepackage{amssymb}
\usepackage[version=4]{mhchem}
\usepackage{stmaryrd}
\usepackage{bbold}
\usepackage{graphicx}

\begin{document}

\title{ {\huge ActPC-Chem: }\\ {\large Discrete Active Predictive Coding \\  for Goal-Guided Algorithmic Chemistry \\ as a Potential Cognitive Kernel \\ for Hyperon \& PRIMUS-Based AGI \\ -  \\ PRELIMINARY DRAFT VERSION}}

\author{Ben Goertzel  \footnote{SingularityNET, TrueAGI, OpenCog } \\
}





\maketitle

\begin{abstract}
This exploratory, speculative "concept paper" explores a novel paradigm (labeled ActPC-Chem) for biologically inspired, goal-guided artificial intelligence (AI) centered on a form of Discrete Active Predictive Coding (ActPC) operating within an algorithmic chemistry of rewrite rules. 

The central thesis of the ActPC-Chem approach is that general-intelligence-capable cognitive structures and dynamics can emerge in a system where both data and models are represented as evolving patterns of metagraph rewrite rules, and where prediction errors, intrinsic and extrinsic rewards, and semantic constraints guide the continual reorganization and refinement of these rules.  

In contrast to backpropagation-based approaches to training large AI networks, ActPC-Chem makes it relatively straightforward to integrate subsymbolic pattern recognition and behavior learning with  symbolic and causal reasoning in a unified framework.

We begin with a review of active predictive coding concepts and show how they can be adapted to a discrete, rewrite-rule-based "algorithmic chemistry" that supports goal-driven reinforcement. To accelerate the evolution of discrete ActPC, we introduce the notion of discrete natural gradients grounded in optimal transport geometry. 

Using a virtual "robot bug" thought experiment, we illustrate how such a system might self-organize to handle challenging tasks involving delayed and context-dependent rewards, integrating causal rule inference (AIRIS) and probabilistic logical abstraction (PLN) to discover and exploit conceptual patterns and causal constraints.

Next, we describe how continuous predictive coding neural networks, which excel at handling noisy sensory data and motor control signals, can be coherently merged with the discrete ActPC substrate. Prediction errors can flow across levels -- continuous sensorimotor layers at the bottom and discrete symbolic reasoning layers at the top -- ensuring stable perception, efficient action control, and logically coherent higher-level cognition.

Finally, we outline how these ideas might be extended to create a transformer-like architecture that foregoes traditional backpropagation in favor of rule-based transformations guided by ActPC. This layered architecture, supplemented with AIRIS and PLN, promises structured, multimodal, and logically consistent next-token predictions and narrative sequences. 

ActPC-Chem is envisioned as a foundational "cognitive kernel" for advanced cognitive architectures, such as the OpenCog Hyperon system, incorporating essential elements of the PRIMUS cognitive architecture. By adding further AI mechanisms from PRIMUS and related frameworks onto ActPC-Chem, this kernel can be expanded into increasingly general and powerful systems, pointing toward a plausible pathway toward human-level artificial general intelligence (AGI) which can then self-modify on to superintelligence (ASI).
\end{abstract}
\newpage
\tableofcontents
\newpage
\section{Introduction}

These are unique times in the history of AI; AGI is finally taken seriously as a scientific and engineering pursuit, and significant resources are now being put into the quest to create AGI and even superintelligence, but there is still nothing near a consensus of which approaches are likely to get us there.

Large neural networks -- trained with massive datasets via backpropagation -- have achieved impressive successes across domains such as language modeling, computer vision, and game-playing. Yet from an GI perspective, these models remain limited (see \cite{goertzel2023LLM} for a detailed exposition on these limitations).  They struggle to exhibit robust causal reasoning, often produce logically inconsistent "hallucinations" and rely on a batch-mode, non-localized training methodology that in many ways contradicts basic principles of adaptive cognition.   Moreover, working around these shortcomings via hybridizing these neural nets with other AI tools is highly challenging -- the current paradigm makes it infeasible to seamlessly integrate symbolic knowledge, abstract reasoning or evolutionary creativity with the neural learning process.   These various issues are not intrinsic to the neural net paradigm generally speaking but have to do with the specific (not very biologically realistic, as it happens) neural net architectures and training approaches currently in use.

On the other hand, purely symbolic or logic-based AI approaches have offered interpretability and direct representations of causality, but historically have failed to scale up and adapt swiftly to the noisy complexity of real-world tasks.   Symbolic evolutionary methods have demonstrated impressive creativity, but also have not been scaled to the level needed for widespread practical utility.   

The longstanding dichotomy between subsymbolic pattern-learning and symbolic abstraction is still with us today -- our large subsymbolic networks, while in some ways surprisingly good at reasoning, remain unable to extrapolate very far beyond their training data; and our symbolic reasoning systems, while in some cases very powerful, have not yet demonstrated the large-scale pattern recognition and synthesis abilities of the best large neural networks.

A key challenge, then, is to discover an architectural principle that naturally blends the various aspects of human-like general intelligence: subsymbolic adaptability, symbolic and causal reasoning, evolutionary creativity and robust experiential learning. 

The PRIMUS cognitive architecture \cite{goertzel2023hyperon}, which partially motivated the OpenCog Hyperon AGI infrastructure, provides a complex and coherent proposal regarding how to overcome these challenges and create a human-level AGI capable of self-modifying itself into an ASI.   However, PRIMUS is a quite large and complicated design, with many aspects only partially specified, and in setting about specifying and implementing PRIMUS in practice there is no single clear best way to start.

Here we propose a specific approach to experiential learning for AGI systems called ActPC-Chem, which makes sense on its own, but also constitutes a subset of the PRIMUS architecture.  In a PRIMUS context, we suggest, ActPC-Chem may be considered as one of multiple possible "cognitive kernels" -- autonomous and reasonably powerful learning algorithms that can control intelligent agents on their own, but can also serve as hubs onto which the other parts of PRIMUS can be attached, either all at once or incrementally.

While PRIMUS as a whole is more cognitively than biologically inspired, ActPC-Chem draws heavily on both biology and chemistry.   

The ActPC part of the name stands for Active Predictive Coding, which is the application of Predictive Coding based learning in a reinforcement learning like context.   Continuous predictive coding models, such as those of Alex Ororbia \cite{ororbia2022neural} which serve as inspiration for much of the design proposed here, act on the premise that intelligent agents continuously predict their inputs and minimize surprise.  Our contribution in this domain is to suggest ways of applying these Active PC learning principles to more discrete and symbolic domains, especially in the context of self-organizing networks of rewrite rules.

The Chem part of the name refers to "algorithmic chemistry" -- the creation of "digital primordial soups" of small software codelets that transform perception inputs and generate output actions, but also transform each other.   One has codelets rewriting codelets into other codelets in a complex network of self-transformation.   This idea goes back decades \cite{fontana1990algorithmic} and has been the subject of slow-paced but ongoing research activity \cite{lightfield2021logics}, and in small experiments it has demonstrated considerable power for self-organization and computational creativity, but it has never been deployed at anywhere near the scale of modern neural networks.  The closest thing to a scalable deployment of algorithm chemistry has probably been some of the experiments with the AERA cognitive architecture \cite{thorisson2020seed}, though AERA also has various special characteristics distinguishing it from typical algorithmic chemistry.

Here, we bring ActPC and algorithmic chemistry together for the first time, and propose Discrete Active Predictive Coding for Goal-Guided Algorithmic Chemistry (ActPC-Chem), intended as a "cognitive kernel" for AGI-ish cognitive architectures such as OpenCog Hyperon, and as a substrate into which additional components from the PRIMUS cognitive framework can be integrated. 

ActPC-Chem envisions a fluid "algorithmic chemistry" of rewrite rules: a metagraph in which both data and models are represented as continually evolving metagraph transformation patterns, which ongoingly rewrite and transform {\it each other} as well as external data in a sort of digital primordial soup.   The weights on these rewrite rules are then updated via a discrete version of Predictive Coding based learning, providing a roughly RL style framework for guiding the self-organizing evolution of the "rewrite rule primordial soup" toward a state that provides utility to the agent in whose mind it lives.

This approach naturally lends itself to machine creativity.  An algorithmic chemistry is, by its nature, an autopoietic network where new patterns spontaneously emerge, combine, and compete. This provides a rich ground for novelty generation and the serendipitous discovery of new strategies and concepts.   By combining such a generative, inventive substrate with predictive coding principles, we can channel this intrinsic creativity along pathways that reduce prediction error and achieve specific goals. 

In the ActPC-Chem approach, reinforcement learning (in a broad sense, including epistemic and instrumental rewards) is used to guide which "molecular" rewrite-rule reactions persist and which vanish. Meanwhile, higher-level symbolic and causal reasoning layers may be introduced via integrating additional mechanisms like AIRIS and PLN, and ultimately the full range of cognitive methods in the PRIMUS architecture (and more).   Such methods can shape and refine the emergent creativity of the network, ensuring that the system not only produces new patterns but does so in a manner aligned with logical coherence and causal correctness. The result is (we conjecture) a framework that integrates the strengths of evolutionary, self-organizing processes with goal-driven, reward-modulated learning and symbolic logical structure.

\subsubsection{Plan of the Paper}

We start by reviewing the concepts of active predictive coding and showing how they can be adapted to discrete, rewrite-rule-based algorithmic chemistries that support goal-driven behavior and reinforcement signals. Next, to accelerate the evolution of discrete ActPC, we discuss the use of discrete natural gradients derived from optimal transport geometry, providing a more stable and geometry-aware update mechanism. Through a "virtual robot bug" thought experiment, we demonstrate how the system might handle complex tasks involving delayed and context-dependent rewards. Integrating causal rule inference (AIRIS) and probabilistic logical abstractions (PLN) enables the discovery of subtle conceptual patterns and causal constraints that guide the system's autopoietic creativity toward effective, adaptive solutions.

We then describe how continuous predictive coding neural networks can be merged with this discrete substrate, creating a hierarchical architecture where continuous sensorimotor loops provide stable perceptions and actions at the bottom, and discrete symbolic reasoning handles abstraction and causality at the top. This synergy allows prediction errors to propagate smoothly across levels, yielding a coherent multimodal pipeline that can continuously refine both perceptual accuracy and strategic intelligence.

Finally, we outline how these ideas could be extended to form a transformer-like model that forgoes formal neurons and backprop in favor of weighted rewrite rules and ActPC-driven rule transformations. The resulting layered architecture --supplemented by AIRIS's causal logic, PLN's abstractions, and continuous PC's robust multimodal integration -- appears well suited to produce structured, contextually rich next-token predictions. 

Summing up in a general way: Beyond the various technical novelties, the broader vision here is that by grounding AGI-capable cognitive processes in an autopoietic, algorithmic-chemistry-like substrate guided by predictive coding, we achieve a system with intrinsic flexibility, adaptability, and inventive power at its core. This "cognitive kernel" may then serve as the base upon which further PRIMUS methods and other AI techniques can be layered, ultimately paving a plausible pathway toward human-level AGI and beyond.

\subsection{ActPC-Chem as one more PRIMUS Probabilization}

A more PRIMUS-related way to summarize these ideas and how they fit into the PRIMUS architecture might be: ActPC-Chem is the next in a series of "probabilizations" of AI methods that has been carried out in an effort to leverage probabilistic semantics to more effectively connect diverse techniques with varying mathematical and conceptual underpinnings into a coherent and synergetic framework.

Predecessors in this vein have been

\begin{itemize}
\item PLN (Probabilistic Logic Networks), which makes highly general and abstract forms of logic probabilistic, in a way that helps connect them to observational data
\item MOSES (Meta-Optimizing Semantic Evolutionary Search), which brings (probabilistic) Estimation of Distribution Algorithms (EDAs) to evolutionary program learning in a general and AGI-friendly way
\item ECAN (Economic Attention Allocation), which can be viewed as a variation of attractor neural networks in which (the analogue of) activations sum to 1, thus making them straightforwardly model-able as probabilities
\end{itemize}

Ororbia's Predictive Coding approach has many interesting specificities to it, but at a high level, it also has the general aspect of making the learning operations within a neural network clearly probabilistically interpretable, much more than is the case with backpropagation (prediction errors are measured via entropies which are transformed probabilities).

Algorithmic chemistry and related approaches have been discussed in a PRIMUS and OpenCog context for decades now, however ActPC-Chem is the first time they have been connected in a clear way with a probabilistic semantics.  It has been proposed previously to use PLN and MOSES to reason probabilistically about which patterns in algorithmic-chemistry networks seemed to be more effective (part of the "Cogistry" proposal \cite{goertzel2016cogistry}), but this is different than using probabilistic methods as the core base-level method of "training" an algorithmic chemistry network, which is what occurs in ActPC-Chem.

There are some echoes here of prior work combining the (algorithmic chemistry like) AERA architecture with Pei Wang's NARS uncertain reasoning engine \cite{rorbeck2022self}.   However, the lack of probabilistic semantics in NARS (among other factors) meant that the math of algorithmic-chemistry-network learning and the math of probabilistic reasoning could not be as tightly connected within that system, as what we believe can be achieved using a common probabilistic semantics in an ActPC-Chem context.

\subsubsection{Probabilistic Graph Structured Lambda Theory as Potential Initial Cognitive-Synergetic Glue}

Having multiple components of a tightly-coupled hybrid cognitive architecture (like PRIMUS) all use probabilistic semantics is a good start toward having the components all "speak the same language" in a manner that's useful for cooperative problem solving ("cognitive synergy").   However it doesn't go far enough.

Recent work on the MeTTa language (the AGI scripting language of the OpenCog Hyperon system in which PRIMUS is being implemented) and the semantics of PLN has led to the tentative conclusion that a specific logic formulation called "Graph structured lambda theory" (GSLT) may be appropriate and sufficient to take us a long way toward cognitive-synergetic AGI.   There are also clear ways to develop probabilistic semantics for GSLT, so that the probability-as-cognitive-glue and GSLT-as-cognitive-glue concepts can work together very closely.

Conceptually, we believe AGI frameworks like Hyperon and PRIMUS should be viewed as going beyond any particular mathematical formalisms, and as they advance in general intelligence they must be able to adapt and conceive their own foundations and formalisms rather than being stuck with those imposed by their human creators.   However, having the right seed formalisms is also important, and our current hypothesis is that probabilistic GSLT may be a highly effective initial formalism to use for cross-connection of the multiple diverse cognitive modules of a tightly-integrated hybrid AGI system like PRIMUS.

Algorithmic chemistry involves a choice of what programming framework to use for the "algorithms" that are chemically combining with each other; in a Hyperon/PRIMUS context these are most naturally taken to be rewrite rules, and that is the avenue developed here in ActPC-Chem.   However, there can be various sorts of rewrite rules with widely varying levels of generality.   In the simplistic concrete examples given here we will use very simple rewrite rules, but our intuition is that as we experiment with these systems, we will probably want more abstract rules than the ones in these examples.   We will probably end up with algorithmic chemistry codelets at some level of abstraction inbetween the very simplistic examples given here, and the full power of GSLT.   Within the overall ActPC-Chem framework outlined here, there is quite a lot to be experimented with.

\section{ActPC-Chem: Discrete Active Predictive Coding for Goal-Guided Algorithmic Chemistry}

Let us review what ActPC is, and how it may be adapted to a discrete setting, and in particular a "metagraph rewrite rule soup" style algorithmic-chemistry setting.

\subsection{Brief Review of Active Predictive Coding}

Active Predictive Coding (ActPC) \cite{ororbia2022neural}  is a novel reinforcement learning (RL) paradigm grounded in predictive coding principles and implemented through Neural Generative Coding (NGC) circuits. Predictive coding, a biologically inspired theoretical framework, posits that the brain continuously predicts sensory inputs and updates internal representations by reducing prediction errors.   Some inspiration has been drawn here from the conceptual neuroscience models of Karl Friston \cite{friston2009free}, however the technical and mathematical underpinnings of contemporary ActPC work are significantly different from any of Friston's specific published proposals.

ActPC leverages the predictive-coding concept to guide action selection and learning without relying on backpropagation-based gradient computations. Instead, it uses local, Hebbian-like learning rules to update synaptic weights, making the approach both biologically plausible and computationally robust (yet far more efficient than simplistic Hebbian learning heuristics).

Key high level motivations of the ActPC approach include:

\begin{enumerate}
  \item Biological Plausibility: Traditional RL methods often depend on backpropagation, a mechanism not readily aligned with known neural processes. ActPC uses local error signals and Hebbian updates, closely mirroring how real neurons might learn.
  \item Sparse Rewards: Many RL tasks, particularly in robotics, provide rewards infrequently. Conventional backpropagation-based RL methods struggle here due to unstable gradient signals. ActPC mitigates this by incorporating both exploratory (epistemic) and goal-oriented (instrumental) signals, enabling effective learning in environments where explicit rewards are rare.
  \item Gradient-Free Optimization: By entirely sidestepping backpropagation, ActPC is not affected by issues like vanishing gradients and differentiability constraints. This is especially useful when employing complex, biologically realistic neuron models or when scaling to large networks.
\end{enumerate}

\subsubsection{Neural Generative Coding (NGC) Framework:}

Ororbia's NGC framework is a specific instantiation of the ActPC idea, with the following key aspects:

\begin{itemize}
  \item Core Idea: NGC layers predict the activity of subsequent layers. The difference between actual and predicted activity generates an error signal, which drives synaptic updates.
  \item State Updates: Neuron states are iteratively updated based on error signals and local connectivity. A layer's activity, $z^{\ell}$, is refined through a dynamic inference process where the error neurons $e^{\ell}=z^{\ell}-\hat{z}^{\ell}$ guide adjustments.
  \item Predictions: Predictions $\hat{z}^{\ell}$ are generated from forward synaptic connections and memory terms $m_{t}$, reflecting temporal context.
  \item Hebbian-Like Updates: Synaptic weights $W$ and error synapses $E$ are adjusted locally using Hebbian-like rules that strengthen connections contributing to accurate predictions.
\end{itemize}

ActPC, as manifested in NGC, integrates multiple predictive circuits to jointly solve the RL problem. Two key signals drive behavior:

\begin{itemize}
  \item Epistemic Signal (Exploration): Encourages the agent to maximize prediction errors, effectively probing uncertain states and learning from them. This forms an intrinsic reward proportional to the squared prediction errors.
  \item Instrumental Signal (Goal-Directed Behavior): Encourages the agent to minimize prediction errors related to goal achievement. This aligns with traditional RL's objective of maximizing cumulative reward.
  \end{itemize}
  
\noindent A combined reward function fuses these signals, balancing exploration and exploitation to achieve stable and robust policy learning.  

This reward function drives action and reinforcement dynamics in a manner roughly similar to classical RL based systems:

\begin{itemize}
  \item Action Model: Motor outputs $a_{t}$ are produced by an NGC-inspired motor-action circuit, with actions bounded and generated via tanh-based nonlinearities.
  \item Policy Estimation: A policy circuit estimates expected returns $q_{t}$. Target values incorporate immediate rewards and future value estimates, analogous to standard RL value functions.
\end{itemize}

Intended advantages of this framework over backpropagation-based RL include:

\begin{itemize}
  \item Biological Plausibility: Local error-driven synaptic updates align with how the brain might learn, eliminating the need for global gradient signals.
  \item Sparse Reward Robustness: The combined reward structure (epistemic + instrumental) ensures that learning progresses even when external rewards are rare.
  \item Gradient-Free: Without backpropagation, ActPC avoids vanishing gradients and is more compatible with diverse neuron and activation models.
  \item Dynamic Adaptation: The iterative inference and updating process allows continuous adaptation in nonstationary environments.
\end{itemize}

Experiments reported in publications from Ororbia's lab validate these advantages to a significant extent, but they have not yet been demonstrated in large-scale practical or commercial applications.  However it seems a promising direction to explore that, via blending exploration-driven epistemic signals with goal-oriented instrumental signals in ActPC, one can achieve stable, adaptive, and efficient learning at a level of capability beyond conventional RL based approaches. 

\subsection{Toward Discrete ActPC: General Considerations}

While the standard math of ActPC is continuous-variable, there is nothing in the conceptual underpinnings of the method that especially favors continuous over discrete implementation.   For applications more toward the cognitive than perception or actuation side, discrete versions of ActPC might have advantages, in terms of naturalness of handling symbolic and abstract knowledge.   Discrete and continuous versions of ActPC would then have natural pathways for close coordination due to their common foundation.

To make a discrete analogue of ActPC, one might adapt the underlying principles of predictive coding and error-driven, local learning rules to a symbolic domain where programs, rather than neural activations, serve as the generative models. This discrete framework would learn to produce or refine programs so that their outputs match observed data or achieve desired goals. Instead of using continuous error neurons and Hebbian weight adjustments, this version would rely on the manipulation of discrete structures -- program instructions, rewriting rules, combinational logic -- and measure error using information-theoretic quantities.

\subsubsection{ Moving Toward a Discrete Variation}

One direction for cashing out this idea might be:

\begin{itemize}
\item{Programs as Generative Models:} In the neural version of ActPC, each layer predicts the activity of downstream layers. In a discrete setting, imagine a program (e.g. a functional program, a logic program, or a sequence of instructions) attempting to generate or predict the observed data from an environment. Here, the program's output is compared to the actual observed output (or target state) at each step. The program thus serves as a generative model, trying to "explain" the incoming data.

\item{Measuring Error via Information Theory:}.  Instead of using the difference between predicted and actual neuron activations, we can measure the "error" as the discrepancy in information content. There are two clearly promising approaches here:

\begin{itemize}
  \item Shannon Information: Use measures like cross-entropy or Kullback-Leibler (KL) divergence between the predicted probability distribution over possible outcomes and the actual observed outcome distribution. The closer the predicted distribution is to the actual one, the less "surprise" or "error."
  \item Algorithmic Information/Complexity: More ambitiously, one could approximate the algorithmic complexity (Kolmogorov complexity) difference between the program's predicted output and the observed output. For instance, if the observed data can be succinctly generated by the program's current form, the complexity is low, implying low error. If the observed data is more complex relative to the program's predictions, the discrepancy is high, and so is the "error."
\end{itemize}

\noindent While exact Kolmogorov complexity is not computable, practical approximations (e.g. via compression-based heuristics or learned generative models) could serve as a complexity-sensitive error measure.

\item{ Local Rewriting Rules as "Synaptic" Updates:}. Neural ActPC uses local, Hebbian-like learning rules to adjust weights. In the discrete setting, we can define a set of local rewriting rules or program transformations that serve as the analogue of weight updates. For example:

\begin{itemize}
  \item Replace an instruction with another that reduces the complexity of explaining the current data.
  \item Insert a conditional branch that better partitions the input cases, reducing overall surprise.
  \item Remove or generalize certain code segments to improve predictive power or reduce unnecessary complexity.
\end{itemize}

\noindent Each rewrite is guided by whether it reduces the measured information-theoretic error. Thus, these local code edits play the role of synaptic adjustments, but in a symbolic domain.

\end{itemize}

\subsubsection{ Epistemic vs. Instrumental Signals in a Discrete Domain}

In ActPC, learning is driven by two complementary signals: an epistemic (exploration) signal that rewards high prediction error (surprise), encouraging the model to seek new informative states, and an instrumental (goal-oriented) signal that rewards states where the prediction error related to a desired goal is minimized.

Similarly, in a program-learning context, the epistemic signal can encourage the system to explore programs that increase its ability to compress or predict previously unpredictable data. High epistemic reward might come from discovering shorter or more elegant programs that generate a good internal model of the environment. In an information-theoretic sense, this can be related to seeking out complexity that can later be compressed: the system is incentivized to explore patterns it cannot currently explain well, thereby gathering more information.

The instrumental reward can be tied to how closely the program's final output matches a desired criterion (e.g., solving a puzzle, reaching a target state, or producing a successful control sequence). Reducing this goal-related complexity difference (or prediction error) provides a strong incentive to refine the program's logic so that it consistently produces the correct outcome.

The combined objective would blend these signals. The system tries to balance improving predictive coverage of the environment (epistemic) with achieving task success (instrumental).

\subsubsection{ Incremental Learning for Discrete ActPC}

Just as the neural ActPC avoids gradient based updates, this discrete analogue would also be gradient-free. Instead of gradients, it uses local transformations guided by whether they reduce the chosen measure of error. This could be implemented as a form of hill-climbing or stochastic search over a space of programs, e.g.:

\begin{itemize}
  \item From a given program, propose local rewrites (akin to random mutations).
  \item Evaluate the new program's predictive error (using Shannon or approximate algorithmic measures).
  \item Accept rewrites that reduce error and reject those that increase it, thus performing a form of local, error-driven adaptation.
\end{itemize}

\noindent However there are many ways to organize this kinds of search, which may have major influences on the efficiency and capability of the resultant framework.

\subsubsection{Choices!}

Summarizing based on the above conceptual considerations, there are many choices to be made in crafting a discrete analogue of classical ActPC, e.g.

\begin{itemize}
\item Choice of Information Measure:

\begin{itemize}
\item Shannon-based measures (like KL divergence) are more tractable and can be computed if you can model the environment's probability distributions.
\item Approximating algorithmic complexity requires compression-based metrics or other heuristics, which may be computationally expensive but could yield more fundamental insights.
\end{itemize}

\item Program Representation: Representing programs in a way that supports local, meaning-preserving transformations is crucial. Functional or logic programming languages, or graph-based intermediate representations, might lend themselves well to well-defined local rewrites.

\item Search and Optimization Methods:. Techniques like genetic programming, Markov chain Monte Carlo (MCMC), or other search heuristics could be employed to navigate the space of programs. The difference is that your "fitness" function (akin to negative error) now explicitly encodes predictive power and compressive capacity, rather than just a task-specific numerical reward.
\end{itemize}

\subsection{ActPC-Chem: Discrete ActPC for Goal-Guided Algorithmic Chemistry}

Now we will get more specific, and propose a particular flavor of discrete ActPC in the spirit of the above general considerations.

Let's consider a self-referential "algorithmic chemistry" of metagraph rewriting rules, where the entire AI system-its data, its models, and the rules used for rewriting-are all encoded as subgraphs within a larger metagraph.  This sort of system would be extremely natural to implement in the MeTTa language \cite{meredith2023metta} of the OpenCog Hyperon framework , and has been discussed as part of the PRIMUS architecture, motivated largely by its potential for radical computational creativity.

In this framework, one starts with:

\begin{itemize}
  \item Metagraph: A graph of graphs, or more generally, a network of nodes and edges where nodes and edges can themselves be graphs. Some parts of this metagraph correspond to "data" (inputs, outputs, intermediate states), while other parts correspond to "code" or "rules" (rewrite rules that operate on the metagraph).
  \item Rewrite Rules: These are transformations that take a certain sub-metagraph pattern (input pattern) and produce a new sub-metagraph pattern (output pattern). Each rewrite rule can be represented as a pair: a "condition pattern" and a "replacement pattern." Because rewrite rules themselves are sub-metagraphs, the system can modify and rewrite its own rules. This enables a fully self-referential system where learning and adaptation emerge from rewriting the rewrite rules.
  \item Inputs and Outputs: Some nodes in the metagraph are designated as "external inputs" originating from the environment (e.g., sensor readings), and some nodes are designated as "external outputs" that affect the environment (e.g., actuator commands). The system applies its rewrite rules iteratively to propagate and transform input patterns into output patterns, effectively computing actions or predictions.
\end{itemize}

To translate the core features of ActPC into a metagraph rewriting context, we consider:

\begin{itemize}
\item  Generative Modeling via Rewrite Rules: The system's current set of rewrite rules constitutes its generative model of how inputs should transform into outputs. If the system is intended to predict future states of the environment or determine the best actions given inputs, its rewrite rules encode these transformations. Applying the rules to the input sub-metagraph ideally should produce sub-metagraphs that match observed or desired outputs.  
\item Collective Rewrite Rule Activity is Critical:  Key to understand here is that we are viewing the rewrite rules as atomic elements of large "virtual chemical solutions" -- so it's the statistical properties of the outputs of a large number of rewrite rules that are important, not any single output of any particular rule.  In this sense we are thinking of the rewrite rules vaguely similarly to neurons in a neural network, which each individual neuron is generally of limited significance and the collective result of the whole neural network is the key thing.
\item Errors and Information-Theoretic Measures: Instead of computing a numeric difference (as in neural net predictive coding), we define an error measure in terms of the divergence between what the rewrite rules generate and what is actually observed. Two possible measures:

\begin{itemize}
\item Shannon Information Measures: Compute something like the cross-entropy or KL divergence between the probability distributions implied by the rewrite rules and the actual observed structures in the metagraph. If the system's rewrite rules, when stochastically applied, predict a certain pattern of outputs, and the observed outputs differ, the resulting "surprise" or information gain is our error.
\item Approximate Algorithmic Information Measures: More ambitiously, approximate the algorithmic complexity (e.g., via compression-based methods) of describing the observed data relative to the generative rewriting rules. If the current rule set succinctly "compresses" and reproduces the observed patterns, error is low. If the observed patterns are more complex than what the current rule set can easily generate, the discrepancy indicates higher complexity and thus higher error.
\end{itemize}

\noindent One way or another, this error acts as the "prediction error" signal analogous to the difference between predicted and actual neural activations in PC.

\item Local Learning via Rule Rewrites: In neural ActPC, weights are adjusted locally based on error signals. In the metagraph scenario, learning corresponds to locally rewriting the rewrite rules themselves. Each rule is represented as a sub-metagraph; to learn, we propose modifications to these sub-metagraphs:

\begin{itemize}
  \item Add, remove, or generalize pattern matches.
  \item Simplify or elaborate the replacement patterns.
  \item Introduce conditional structures that branch differently based on detected subpatterns.
\end{itemize}

\item Each candidate modification (rewrite of a rewrite rule) is evaluated by applying the modified rules to input states and measuring the new information-theoretic error. If the change reduces error, we retain it; otherwise, we reject or revert it. This is a local, gradient-free update analogous to Hebbian-like synaptic changes, but now operating on discrete structures.

\end{itemize}

\subsubsection{Epistemic and Instrumental Signals}

The two types of reward signals would behave familiarly in this setting:

\begin{itemize}
\item Epistemic Signal (Exploration): The system is rewarded for discovering new rewrite rules that reduce its uncertainty or surprise about the environment. If the environment data is currently not wellcompressed by the model, introducing rewrite rules that better "explain" or "compress" these patterns yields a positive epistemic reward. This encourages the system to explore more complex or detailed rules to handle new patterns.
\item Instrumental Signal (Goal Achievement): Some output patterns correspond to achieving a goal (e.g., producing the correct actuator commands to solve a robotic task). The system receives an instrumental reward for rewriting rules that reduce the complexity/distance between predicted outputs and desired outputs. For example, if the environment's input-output mapping is more easily generated by a simpler set of rewrite rules after a change, that's a gain in instrumental reward.
\end{itemize}

One can think of the entire system as a chemical soup of rewrite rules, where rules "react" with the metagraph and with each other. The "reactions" that survive and propagate are those that yield lower information-theoretic error and higher combined rewards. Over time:

\begin{itemize}
  \item Rules that fail to reduce error or help achieve goals are effectively "extinguished" or replaced.
  \item Rules that reduce complexity, improve predictive accuracy, or achieve better outcomes stabilize and become more prevalent.
\end{itemize}

This creates a self-organizing process, an "algorithmic chemistry" of meta-rules evolving toward better predictive and goal-achieving capability.

\subsubsection{Self-Modification and Bootstrapping}

Because rewrite rules can rewrite other rewrite rules, the system can develop meta-level strategies. For example, it can invent "metarules" that:

\begin{itemize}
  \item Identify overly complex or uncompressed regions of the metagraph.
  \item Rewrite those rules in ways known to reduce complexity.
  \item Introduce structured macros or higher-level abstractions that simplify large portions of the metagraph.
\end{itemize}

\noindent  The concept is that, over time, such bootstrapping leads to a hierarchical structure of models-akin to how predictive coding hierarchies form in neural networks. The system might evolve layers of rewriting rules, where top-level meta-rules govern the structure and complexity of lower-level rules, guiding them to form simpler, more coherent generative explanations of the environment.

This framework is what we call {\bf ActPC-Chem} -- discrete ActPC applied to rewrite-rule algorithmic chemistry.

\subsubsection{Similarities to AERA}

Kristinn Thorisson's AERA (Autocatalytic Endogenous Reflective Architecture) architecture \cite{thorisson2020seed} involves self-programming, modular models that predict and learn incrementally.  The proposed ActPC-Chem design  dovetails with this idea in multiple ways:

\begin{itemize}
  \item In AERA, models predict future events and adjust themselves to improve predictions. Similarly, in the algorithmic chemistry scenario, rewrite rules continuously evolve to better predict and generate target patterns.
  \item The environment, input patterns, and desired outcomes form constraints on the rewriting process. The system's emergent codebase (the set of rewrite rules) is always under pressure to become more compressive and more effective.
  \item Much like AERA's self-reflective modeling, the metagraph rewriting system is reflective: it treats its own rules as data to be improved upon. The loop of predict $\rightarrow$ measure error $\rightarrow$ locally rewrite rules $\rightarrow$ improve predictions creates a dynamic, incremental learning process without explicit gradients.
\end{itemize}

AERA, however, makes some specific commitments regarding rewrite rule representation, execution, learning and attention which are not necessary for the general concept of "ActPC on rewrite rules", and which don't always gel naturally with the mathematics of NPC nor with other tools within PRIMUS/Hyperon.

\subsection{Discrete ActPC for Goal-Guided Algorithmic Chemistry: A Naive Formalization}

We will now elaborate one possible way to more precisely formulate an "ActPC-Chem" system -- a system of equations analogous to Ororbia's ActPC, but for a discrete, algorithmic-chemistry style metagraph rewriting system with an information-theoretic error measure. The system is more abstract and involves nondifferentiable, discrete updates, but we can still write down analogous mathematical relationships.

We will start, here, with a somewhat algorithmically naive (and perhaps overly inefficient) approach, then in later sections make it more sophisticated via introducing appropriate notions of gradient on probability distributions over discrete sample spaces.

\subsubsection{Notation and Setup}

Consider the following setup:

\begin{itemize}
  \item Metagraph: At time step $t$, we have a metagraph $G_{t}$. This metagraph includes, among other portions, two two critical sub-metagraphs:
  \begin{itemize}
  \item Input Subgraph $G_{t}^{i n}$ : Nodes/edges designated as external inputs from the environment at time $t$.
  \item Output Subgraph $G_{t}^{\text {out }}$ : Nodes/edges designated as external outputs that the system produces.
  \end{itemize}
  \item Rewrite Rules: The system maintains a set of rewrite rules $R_{t}=\left\{r_{i}\right\}_{i=1}^{N}$. Each rule $r_{i}$ is itself a sub-metagraph that specifies how to transform one sub-pattern into another:

$$
r_{i}: P_{i}^{(i n)} \rightarrow P_{i}^{(o u t)}
$$

\noindent where $P_{i}^{(\text {in })}$ and $P_{i}^{(\text {out })}$ are sub-metagraph patterns.
\end{itemize}

The generative process naturally corresponding to this setup is straightforward: Applying the rewrite rules $R_{t}$ to $G_{t}^{i n}$ generates a predicted output structure $\hat{G}_{t}^{\text {out }}$. 

This can be viewed as:

$$
\hat{G}_{t}^{o u t}=\Gamma\left(R_{t}, G_{t}^{i n}\right)
$$

\noindent where $\Gamma$ is the generative operator that repeatedly matches $P_{i}^{(i n)}$ of each applicable $r_{i}$ and rewrites it into $P_{i}^{(o u t)}$, until no further rewrites apply or a termination condition is met.

The environment provides the actual observed output configuration $G_{t}^{\text {out }}$ at time $t$.

\paragraph{Distribution of Patterns}: Assume the system can be stochastic (using a stochastic selection rule for choosing which rules to activate, which would be the natural approach in a PRIMUS context) and/or that we can derive a probability distribution over certain output patterns $m$ based on how often they appear under the rewrite rules.  We can then look at a distribution like:

$$
p_{t}(m)=P\left(m \mid R_{t}, G_{t}^{i n}\right), \quad q_{t}(m)=P\left(m \mid G_{t}^{o u t}\right)
$$

\noindent where $p_{t}(m)$ is the predicted probability (or frequency) of pattern $m$ given the current rewrite rules, and $q_{t}(m)$ is the observed distribution from the actual environment.

This distribution is the key to applying ActPC based ideas to the rewrite rule soup.  As noted above, it is critical that we are viewing individual rewrite rules as less essential, and focusing attention on the collective statistical behaviors of the whole rewrite rule population and its subpopulations.

\paragraph{Stochastic Rule Selection}: The most natural way to manage stochastic rule selection in a PRIMUS context would be using some form of ECAN attention allocation to update STI (short-term importance) values associated with the Atoms in the Atomspace metagraph representing individual rewrite rules.   Those rules with small prediction error would get an STI stimulus, which would then feed into the overall STI formula to adjust the STI level of the rules.

The probabilistic semantics of ECAN intersects nicely with the probabilistic semantics of ActPC here, in ways that could be elaborated in detail.   Prior experimental work using natural gradients to accelerate ECAN \cite{ikle2011nonlinear} might also end up relevant here, and could be interestingly synergized with ideas to be presented below on natural gradients for accelerating learning in ActPC-Chem. 

\paragraph{Simple Error Measure}: Given this probabilistic setup, we define an information-theoretic error measure. A natural choice is a divergence between the observed and predicted distributions. For instance, we can use the Kullback-Leibler (KL) divergence:

$$
e_{t}=D_{K L}\left(q_{t} \| p_{t}\right)=\sum_{m} q_{t}(m)\left[\log q_{t}(m)-\log p_{t}(m)\right]
$$

\noindent This $e_{t}$ serves as the "prediction error" analogous to the difference between predicted and actual states in neural ActPC.

\subsubsection{Rewards (Epistemic and Instrumental)}
We can define two forms of reward that guide how we modify the rewrite rules:

\begin{itemize}
  \item Instrumental Reward $\left(r_{t}^{i n t}\right)$ : Encourages the system to produce outputs that achieve a certain goal. This is analogous to the traditional RL reward. We can tie it directly to reducing the error:

$$
r_{t}^{i n t}=-e_{t}
$$

  \item Epistemic Reward $\left(r_{t}^{e p}\right)$ : Encourages the system to explore patterns that it does not currently compress or predict well. One way is to reward complexity reduction or "surprise" that leads to better future modeling. For simplicity, let's say the epistemic reward is proportional to the raw complexity or unexpectedness:

$$
r_{t}^{e p}=\sum_{m} q_{t}(m) \log \frac{1}{p_{t}(m)} \quad 
$$

\noindent One could normalize and scale this measure in various ways, but the idea stands that epistemic reward encourages exploration of patterns that are currently poorly predicted.
\end{itemize}

The combined reward can then be:

$$
r_{t}=\alpha_{i n t} r_{t}^{i n t}+\alpha_{e p} r_{t}^{e p}
$$

\subsubsection{Update of Rewrite Rules}
The key difference from neural ActPC is that we do not have differentiable parameters. Instead, we have discrete structures $r_{i}$. Learning means rewriting these rewrite rules to reduce $e_{t}$.

Define a "neighborhood" of small changes to each rewrite rule $r_{i}$. Such changes might include:

\begin{itemize}
  \item Slightly altering $P_{i}^{(\text {in })}$ or $P_{i}^{(o u t)}$.
  \item Adding or removing conditions.
  \item Replacing a sub-pattern with a more general or more specific pattern.
\end{itemize}

Let $\mathcal{N}\left(r_{i}\right)$ be a set of candidate modifications of rule $r_{i}$.   As a crude initial stab, we could attempt to find a new rule $r_{i}^{\prime} \in \mathcal{N}\left(r_{i}\right)$ that reduces the error:

$$
r_{i} \leftarrow \underset{r_{i}^{\prime} \in \mathcal{N}\left(r_{i}\right)}{\arg \min } e_{t}\left(R_{t} \backslash\left\{r_{i}\right\} \cup\left\{r_{i}^{\prime}\right\}\right)
$$

\noindent for each $i$. This is rough, inefficient discrete analogy to a gradient descent step. Instead of computing gradients, we do local search over candidate rule modifications. If none of the modifications reduce the error, we may leave $r_{i}$ unchanged at this step or attempt stochastic replacements.

A little later on, we will explore ways of replacing this with a more efficient approach, leveraging discrete versions of the natural gradient.

\subsubsection{Iterative Inference and Update Cycle}

In the natural core iteration of this kind of system, the steps are:

\begin{enumerate}
  \item Inference (Prediction):

$$
\hat{G}_{t}^{\text {out }}=\Gamma\left(R_{t}, G_{t}^{i n}\right)
$$

\item Error Computation:  Derive distributions $p_{t}(m)$ and $q_{t}(m)$ from $\hat{G}_{t}^{o u t}$ and $G_{t}^{o u t}$, and compute

$$
e_{t}=D_{K L}\left(q_{t} \| p_{t}\right)
$$

\item Compute Rewards:
$$
r_{t}^{i n t}=-e_{t}, \quad r_{t}^{e p}=\sum_{m} q_{t}(m) \log \frac{1}{p_{t}(m)}, \quad r_{t}=\alpha_{i n t} r_{t}^{i n t}+\alpha_{e p} r_{t}^{e p}
$$

\item  Rule Update: As a simplistic initial stab, one might try the following.   For each $r_{i} \in R_{t}$ :

$$
r_{i} \leftarrow \underset{r_{i}^{\prime} \in \mathcal{N}\left(r_{i}\right)}{\arg \min } e_{t}\left(R_{t} \backslash\left\{r_{i}\right\} \cup\left\{r_{i}^{\prime}\right\}\right)
$$

\noindent If such a $r_{i}^{\prime}$ reduces the error, accept it. Otherwise, leave $r_{i}$ as is or consider a probabilistic acceptance criterion.
\end{enumerate}

Over time, these iterative updates change $R_{t}$ to better predict (and thus lower $e_{t}$ ), and to discover more compact, goal-directed rewrite rule sets. This leads to a self-organizing system that refines its generative models (the rewrite rules) for handling the input-output mapping in a manner analogous to predictive coding, but in a discrete, information-theoretic, metagraph rewriting domain.

While these are more schematic procedures than closed-form equations, the structure closely parallels the logic of ActPC:

\begin{itemize}
  \item State and Prediction: $\hat{G}_{t}^{\text {out }}=\Gamma\left(R_{t}, G_{t}^{i n}\right)$
  \item $\quad \operatorname{Error}\left(\right.$ Surprise): $e_{t}=D_{K L}\left(q_{t} \| p_{t}\right)$
  \item Local Updates: $r_{i} \leftarrow \arg \min _{r_{i}^{\prime}} e_{t}$ over local neighborhoods
\end{itemize}

This provides a conceptual and notational framework for extending ActPC-like principles into a fully discrete, self-referential "algorithmic chemistry" of metagraph rewriting rules.

\subsection{A Toy Example}

Toy Scenario: Consider a simple reinforcement learning (RL) problem where an agent must navigate a 1D world (a line of cells) to reach a goal state. The environment is very simple:

\begin{itemize}
  \item Environment: A line of cells numbered from 0 to 3.
  \item Initial State: The agent always starts at cell 0.
  \item Goal State: The goal is cell 3.
  \item Actions: The agent can move either Right (+1) or Left (-1), but since the start is at 0 , initially moving left is either disallowed or just keeps the agent in place if it tries it.
  \item Reward: The agent receives:  $\quad+1$ if it reaches cell 3,  0 otherwise.
\end{itemize}

The problem: The agent needs to figure out a pattern of actions from cell 0 to cell 3 that yields a reward, starting with no prior knowledge.

This is an extremely dumb example and is used here just to give a first notion of how to connect the somewhat abstract ideas and formalisms in the previous section with some specific scenario.

\subsubsection{Representing the System as Rule Rewriting over a Metagraph:}

\begin{enumerate}
\item Metagraph Structure:
\begin{itemize}
  \item We have a metagraph $G_{t}$ at each timestep $t$.
  \item There is a special "input" subgraph $G_{t}^{i n}$ containing:
  \item A node labeled "State" with a value indicating the current cell number.
  \item We want to produce an "output" subgraph $G_{t}^{\text {out }}$ :
  \item A node labeled "Action" with a value "Right" or "Left."
\end{itemize}

\item Rewrite Rules as Generative Models:
\begin{itemize}
  \item We store a set of rewrite rules $R_{t}=\left\{r_{i}\right\}$ that look at the current "State" node and produce an "Action" node.
  \item Example initial rules (totally random guess at the start):
  \begin{itemize}
  \item $r_{1}$ : If you see State $=s$, rewrite it as State $=\mathrm{s}$; Action $=$ Right.
  \item $r_{2}$ : If you see State $=\mathrm{s}$, rewrite it as State $=\mathrm{s}$; Action $=$ Left.
  \end{itemize}
\end{itemize}

\noindent Initially, these rules might be too generic or conflict with each other. The system will try to refine them.

\item Stochastic Application of Rules:

\begin{itemize}
  \item Given the input subgraph with State $=\mathrm{s}$, the system picks a rewrite rule from $R_{t}$ to produce Action.
  \item At the start, imagine it randomly picks actions since it has no preference.
\end{itemize}

\item Observed Outcome and Error: After the agent takes the chosen action, the environment updates the state:

\begin{itemize}
  \item If the agent was at state 0 and took action Right, it moves to 1.
  \item If at state 1 and goes Right, moves to 2; from 2 Right, moves to 3 and gets a reward of +1 .
\end{itemize}

\noindent After the environment moves, we get a new observed "State" node for the next step, and at the end of an episode (or each step), we know if we got a reward.
\end{enumerate}

To measure error, we consider the probability distributions over outcomes implied by our rewrite rules. For simplicity, suppose we define:

\begin{itemize}
  \item $p_{t}(m)$ : The probability distribution over (State, Action) pairs the model predicts based on its rules.
  \item $q_{t}(m)$ : The observed distribution of outcomes after many runs.
\end{itemize}

If we run multiple episodes or steps and compare predicted vs. observed frequencies, we can compute a KL divergence $e_{t}=D_{K L}\left(q_{t} \| p_{t}\right)$. The system tries to minimize this divergence: it wants its rewrite rules to "explain" the successful paths leading to reward.

\subsubsection{ Instrumental and Epistemic Rewards:}

The two kinds of reward may be manifested in this example as follows:

\begin{itemize}
  \item Instrumental Reward: If the agent reaches cell 3, it receives +1 . We tie this to reducing the error between predicted success and actual success. The better the rules lead to reaching cell 3, the lower the error (since the system's internal prediction would converge on a strategy that yields a high probability of correct actions leading to the goal).
  \item Epistemic Reward: If the current rules fail to predict certain transitions (for example, the agent keeps thinking that going left is good but it never leads to reward), there is high surprise. The system is encouraged to find a better set of rewrite rules that "compress" or "explain" the pathway to the goal. This might mean introducing new, more specific rules that map State=0 to Action=Right with higher certainty, reducing the unpredictability in outcomes.
\end{itemize}

Combined:

$$
r_{t}=\alpha_{i n t}\left(-e_{t}\right)+\alpha_{e p}(\text { surprise measure })
$$

\noindent here $e_{t}$ is related to how off the model's predicted distribution is from the observed one. Minimizing $e_{t}$ and balancing exploration (epistemic) vs. exploitation (instrumental) leads the rewrite rules to evolve.\\

\subsubsection{Example Rewrite Rule Updates}

Initially:

$$
\begin{gathered}
r_{1}:(\text { State }=s) \rightarrow(\text { State }=s ; \text { Action }=\text { Right }) \\
r_{2}:(\text { State }=s) \rightarrow(\text { State }=s ; \text { Action }=\text { Left })
\end{gathered}
$$

After this, the  system might find that always going Right from 0 leads it through $1 \rightarrow 2 \rightarrow 3$, and thus to a reward. Over multiple trials, it observes that sequences of Right moves yield high reward. 

To reduce KL divergence (error) and increase instrumental reward, it may then specialize the rules, and e.g. perhaps introduce a new rule $r_{3}$ that says:

$$
r_{3}:(\text { State }=0) \rightarrow(\text { State }=0 ; \text { Action }=\text { Right })
$$

Similarly:

$$
\begin{aligned}
& r_{4}:(\text { State }=1) \rightarrow(\text { State }=1 ; \text { Action }=\text { Right }) \\
& r_{5}:(\text { State }=2) \rightarrow(\text { State }=2 ; \text { Action }=\text { Right })
\end{aligned}
$$

Now, applying these more specialized rules produces a deterministic policy: from $0 \rightarrow$ Right, from $1 \rightarrow$ Right, from $2 \rightarrow$ Right. This yields a guaranteed reward once cell 3 is reached, lowering the KL divergence between predicted and observed outcomes (since the model now "predicts" that following its rewrite rules always leads to success, which matches what is observed).

If the system tries other modifications, such as:

$$
r_{6}:(\text { State }=0) \rightarrow(\text { State }=0 ; \text { Action }=\text { Left })
$$

\noindent it will see that this leads to less reward, higher discrepancy, and thus higher error. Such a rewrite will be rejected or eventually "unlearned."

\subsubsection{ Discrete Update Step:}

Each learning iteration might be:

\begin{enumerate}
  \item Apply current rules to pick an action at the current state.
  \item Execute action, observe next state and whether a reward was obtained.
  \item Estimate $q_{t}(m)$ (the observed pattern distribution) vs. $p_{t}(m)$ (implied by the current rule set).
  \item Compute $e_{t}=D_{K L}\left(q_{t} \| p_{t}\right)$ and rewards $r_{t}^{i n t}, r_{t}^{e p}$.
  \item Propose local changes to rules $r_{i}$ (e.g., specialize a rule to a particular state, remove a bad rule, etc.).
  \item For each proposed change, recalculate $e_{t}$ if that change were made. If $e_{t}$ (or the combined reward) improves, accept the change.
\end{enumerate}

Over iterations, the rules become increasingly aligned with a policy that leads to the goal.

\subsubsection{Example Recap:}

In this toy RL problem, the discrete ActPC-like system:

\begin{itemize}
  \item Starts with generic, uncertain rewrite rules.
  \item Interacts with the environment by applying these rules to produce actions.
  \item Measures how well the predicted outcomes match the observed outcomes and how well it achieves the goal.
  \item Iteratively refines its rewrite rules to minimize information-theoretic error and maximize reward.
  \item Ends up discovering a simple, deterministic rewrite policy mapping states to "Right" actions that lead it from state 0 to state 3, achieving maximum reward with minimal error.
\end{itemize}

This demonstrates how the discrete approach -- treating policies as rewrite rules in a metagraph --can solve a simple RL problem via incremental self-modification informed by information-theoretic errors, much like ActPC does in a neural domain.

\subsection{Incorporating Natural Gradients}

We have presented a concrete instantiation of the general notion of discrete ActPC in the context of rewrite-rule algorithmic chemistry.   However the specific algorithmic approach given above has some obvious inefficiencies to it.  Specifically, the method of search for error-reducing variations of a rewrite rule proposed above is inefficient, involving brute-force search across the neighborhood of the rule.

We will explore here a route to making discrete ActPC-like approaches more efficient by leveraging ideas from discrete optimal transport (DOT) and the Wasserstein natural gradient flow. The aim is to replace a naive local rule-update step with a more geometrically informed optimization step that respects the underlying structure of the space of rewrite rules, potentially leading to more stable and meaningful learning dynamics.

\subsubsection{Context Recap}
In the discrete ActPC scenario, we have:

\begin{itemize}
  \item A set of rewrite rules $R_{t}=\left\{r_{i}\right\}$ defining how input patterns are transformed into output patterns.
  \item A probability distribution $p_{t}$ over certain outcomes or patterns generated by these rules. Alternatively, we can represent the agent's model (its policy or generative structure) as a probability distribution $p_{t}$ over the space of possible rewrite rules or their configurations.
  \item A loss function $F(p)$ that might encode prediction error, complexity, or a combination of epistemic and instrumental factors.
\end{itemize}

The update mechanism involves adjusting rewrite rules to reduce an information-theoretic error measure. Originally, we considered a simple local-search method: propose local modifications and accept those that reduce the error. However, an appealing alternative may be to introduce a DOT-inspired Wasserstein gradient approach to enable search to move in a direction that respects the underlying geometry of rewrite-rule space.

\subsubsection{Incorporating the Wasserstein Natural Gradient}

We now explain how to extend our discrete ActPC using the approach to Wasserstein natural gradient over probability distributions with discrete sample spaces presented in \cite{li2018natural}.

\paragraph{Parameterizing the Distribution Over Rules:}. First, let's define a parameter vector $\xi$ that parameterizes the probability distribution $p(\xi)$ over the set of rewrite rules or rule configurations. For instance, $\xi$ might describe a categorical distribution over different candidate rewrite rules or a set of weights encoding how often certain patterns are used.

The optimization target is to minimize a loss:

$$
F(p(\xi))
$$

\noindent which could be a KL divergence between predicted and observed patterns, or a combined reward objective as in ActPC.

\paragraph{Ground Metric on Rewrite Rules:}.  To apply the Wasserstein gradient, we need a ground metric $\omega_{i j}$ on the discrete space of rewrite rules. Think of each rewrite rule $r_{i}$ as a node in a graph, and $\omega_{i j}$ as the "cost" of transporting probability mass from rule $r_{i}$ to rule $r_{j}$. This metric encodes structural similarities or dissimilarities between rules-for example, how similar the input/output patterns they produce are, or how closely related they are in terms of complexity or function.

Such a graph $G=(V, E)$ over the discrete space of rules allows us to define a weighted Laplacian $L(p)$ and related operators for discrete OT.\\

\paragraph{Wasserstein Geometry and Gradient Flow:} In classical gradient-based updates, we would do:

$$
\xi_{k+1}=\xi_{k}-\eta \nabla_{\xi} F\left(p\left(\xi_{k}\right)\right)
$$

However, this does not consider the geometric structure of the space of probability distributions over rules.

Using the Wasserstein natural gradient flow, we incorporate the geometry defined by the Wasserstein metric. The update rule becomes:

$$
\frac{d \xi}{d t}=-G(\xi)^{-1} \nabla_{\xi} F(p(\xi))
$$

\noindent wwhere $G(\xi)$ is a metric tensor induced by the Wasserstein geometry. Concretely:

\begin{itemize}
  \item Construct the weighted Laplacian $L(p(\xi))$ from the graph and the current distribution $p(\xi)$.
  \item Compute the Jacobian $J_{\xi}$ of the mapping from parameters $\xi$ to the distribution $p(\xi)$
  \item Form:

$$
G(\xi)=J_{\xi}^{T} L(p(\xi))^{\dagger} J_{\xi}
$$

\noindent where $L(p(\xi))^{\dagger}$ is a pseudoinverse of the weighted Laplacian. This ensures that the gradient update takes into account the optimal transport geometry, effectively transporting probability mass in a way that respects the "ground metric" $\omega_{i j}$.
\end{itemize}

\paragraph{Discrete Update Steps (Forward Euler):} A practical update step might be:

$$
\xi_{k+1}=\xi_{k}-h G\left(\xi_{k}\right)^{-1} \nabla_{\xi} F\left(p\left(\xi_{k}\right)\right)
$$

Here:

\begin{itemize}
  \item $\quad \nabla_{\xi} F\left(p\left(\xi_{k}\right)\right)$ is the parameter-space gradient of the loss.
  \item $G\left(\xi_{k}\right)$ is computed from the current distribution and the rule-space geometry.
  \item $h$ is a step size.
\end{itemize}

This update uses the Wasserstein metric to ensure that when we shift probability mass between different rewrite rules, we do so optimally with respect to the defined costs $\omega_{i j}$.\\

\paragraph{Backward Euler (JKO Scheme) for Stability:} For improved stability, especially when dealing with non-smooth or complex loss landscapes, we could use a JKO (Jordan-Kinderlehrer-Otto) scheme:

$$
\xi_{k+1}=\arg \min _{\xi}\left[F(p(\xi))+\frac{\operatorname{Dist}\left(\xi, \xi_{k}\right)^{2}}{2 h}\right]
$$

\noindent where $\operatorname{Dist}\left(\xi, \xi_{k}\right)$ is the Wasserstein distance between $p(\xi)$ and $p\left(\xi_{k}\right)$. This solves a discrete optimal transport problem at each step to find the next parameter configuration that best balances minimizing the loss and staying close (in Wasserstein sense) to the current distribution.

\subsubsection{ Intuition and Benefits}

By incorporating the Wasserstein geometry as suggested, the idea is to:

\begin{itemize}
  \item Instead of making arbitrary local changes to rewrite rules and checking if the error decreases, we get a direction of change that moves probability mass between rules in a globally coherent manner.
  \item If two rewrite rules are "close" in the sense of $\omega_{i j}$, probability mass is more easily shifted between them, leading to smoother adaptation of the model.
  \item The process respects the underlying structure of the rule space, leading to potentially faster convergence and more meaningful intermediate solutions, rather than random jumps in the discrete space.
\end{itemize}

\subsubsection{The Discrete Measure-Dependent Laplacian}

Now we dig a little further into the math of the Wasserstein approach, in particular the key matter of the discrete gradient involved.

The Laplacian in this discrete optimal transport (DOT) setting is a weighted, measure-dependent graph Laplacian constructed from a graph whose nodes represent states (e.g. rewrite rules) and edges encode the "ground metric" between these states. The crucial difference from a standard graph Laplacian is that it incorporates the current probability distribution $p$ over states, making it context-dependent.

\paragraph{Step-by-Step Definition:}
\begin{enumerate}
  \item Nodes and Edges: Consider a finite set of states (for example, the rewrite rules) indexed by $i=1, \ldots, N$. These states form the nodes of a graph $G=(V, E)$.
  \item Ground Metric Weights: Each edge $(i, j) \in E$ is assigned a weight $\omega_{i j}>0$ that encodes the "distance" or "cost" of moving probability mass between states $i$ and $j$. This $\omega_{i j}$ is your ground metric on the discrete space.
  \item Probability Distribution $p$ : At any given time, you have a probability distribution $p=$ $\left(p_{1}, p_{2}, \ldots, p_{N}\right)$ over the nodes (states). The distribution reflects how likely or how much probability mass is currently assigned to each state.
  \item Gradient and Divergence on a Graph: In the discrete setting, the gradient and divergence operators are defined on node-level functions $\Phi: V \rightarrow \mathbb{R}$ and on distributions $p$. For an edge $(i, j)$ , the "gradient" of $\Phi$ is given by:

$$
\nabla_{G} \Phi_{i j}=\omega_{i j}\left(\Phi_{i}-\Phi_{j}\right)
$$

The "divergence" of a vector field (here represented by $p \nabla_{G} \Phi$ ) at node $i$ is:

$$
\left(\operatorname{div}_{G}\left(p \nabla_{G} \Phi\right)\right)_{i}=\sum_{j:(i, j) \in E} \omega_{i j}\left(p_{i}+p_{j}\right)\left(\Phi_{i}-\Phi_{j}\right)
$$

  \item Measure-Dependent Laplacian: The Laplacian $L(p)$ arises as the operator that relates $\Phi$ to $\operatorname{div}_{G}\left(p \nabla_{G} \Phi\right)$. For each node $i$ :

$$
(L(p) \Phi)_{i}=-\left(\operatorname{div}_{G}\left(p \nabla_{G} \Phi\right)\right)_{i}
$$

By substituting the divergence definition, we get:

$$
(L(p) \Phi)_{i}=\sum_{j:(i, j) \in E} \omega_{i j}\left(p_{i}+p_{j}\right)\left(\Phi_{i}-\Phi_{j}\right)
$$

\end{enumerate}

A key point here is that: Unlike a standard graph Laplacian $L=D-W$ that depends only on the graph structure, this operator depends on both the structure ( $\omega_{i j}$ ) and the current distribution $p$. As $p$ changes during the optimization process, the Laplacian $L(p)$ also changes.\\

In optimal transport formulations, the probability distribution $p$ influences the "cost" of moving mass. The measure-dependent Laplacian captures how "resistance" to moving probability mass changes with the current distribution. If $p_{i}$ and $p_{j}$ are large, it may be "easier" to move probability between states $i$ and $j$ because they already hold substantial probability mass; the operator $L(p)$ encodes this dynamic transport geometry.

When computing the Wasserstein natural gradient flow, we invert or pseudoinvert $L(p)$ (or related constructs) to define a Riemannian metric on the space of parameters $\xi$. This gives:

$$
G(\xi)=J_{\xi}^{T} L(p(\xi))^{\dagger} J_{\xi}
$$

\noindent where $J_{\xi}$ is the Jacobian of the parameterization from $\xi$ to $p(\xi)$.

In summary:

\begin{itemize}
  \item You start with a graph encoding state-to-state distances $\omega_{i j}$.
  \item Given a distribution $p$, you build a measure-dependent Laplacian $L(p)$ that incorporates both the graph's structure and the current distribution.
  \item This operator is then used in defining and computing the Wasserstein natural gradients for updating the system's parameters in a way that respects the underlying transport geometry.
\end{itemize}

\subsubsection{ Simplistic Example Scenario:}

Let's flesh this out briefly in a specific example, the very simplistic toy RL problem we discussed above:

\begin{itemize}
  \item Define nodes in a graph representing different candidate rewrite rules (e.g., "From State $=0 \rightarrow$ Action=Right," "From State=1 $\rightarrow$ Action=Left," etc.).
  \item Assign a ground metric $\omega_{i j}$ based on some similarity measure between rules (e.g., how similar the states and actions they use are).
  \item Start with a uniform distribution $p(\xi)$ over rules.
  \item Compute the Wasserstein gradient step using the current distribution and the loss (the KL divergence or reward-based objective).
  \item The update will "transport" probability mass from less successful rules to more promising ones in a principled manner, guided by the geometry of the rule space.
\end{itemize}

Over iterations, the system converges to a distribution of rewrite rules that better solve the RL task, but the process is now smoother and more informed by the structure of the problem, rather than just local ad-hoc modifications.

Overall: By integrating discrete optimal transport ideas into the discrete ActPC framework, we replace naive gradient or local search steps with Wasserstein-based natural gradient flows. This yields a more principled and potentially more efficient update mechanism. The resulting system respects the underlying geometry and relationships between rewrite rules, can incorporate domain knowledge through the choice of the ground metric $\omega_{i j}$, and ultimately should yield more stable and meaningful learning dynamics in discrete, structured RL and program-derivation tasks.

\subsubsection{Recap of ActPC-Chem Process Using Natural Gradient}

To recap the natural gradient approach to ActPC-Chem, then: We represent the distribution over rewrite rules by parameters $\xi$. The rules $R_{t}$ induce $p(\xi)$. To update $\xi$, we use a Wasserstein natural gradient:

$$
\frac{d \xi}{d t}=-G(\xi)^{-1} \nabla_{\xi} F(p(\xi))
$$

\noindent where $F$ could be the negative combined reward (or the error), and $G(\xi)$ is derived from a ground metric on the rule space (via a weighted Laplacian).

The discrete step update looks like:

$$
\xi_{k+1}=\xi_{k}-h G\left(\xi_{k}\right)^{-1} \nabla_{\xi} F\left(p\left(\xi_{k}\right)\right) .
$$

Crude pseudocode suitable for testing out simple versions of this framework this might look like:

\paragraph{Pseudocode}
\paragraph{Initialization:}
\begin{verbatim}
# Initialize metagraph G with input nodes defined
G_in = initialize_input_nodes(env)
G_out = None
# Initialize rewrite rules R
R = initialize_rewrite_rules() # e.g., random or generic rules
# Initialize parameters xi that encode distribution over rules
xi = initialize_params(R)
\end{verbatim}

\paragraph{Training Loop:}

\begin{verbatim}
for iteration in range(max_iterations):
# 1. Generate output by applying rules to input
G_out = apply_rules(G_in, R) # Produces predicted output pattern
predicted_distribution = derive_distribution(G_out, R)
# 2. Observe actual environment outcome and form observed distribution
observed_distribution = derive_observed_distribution(env, G_in)
# 3. Compute error and rewards
e_t = KL_divergence(observed_distribution, predicted_distribution)
r_int = -e_t
r_ep = compute_surprise(observed_distribution, predicted_distribution)
r_t = alpha_int * r_int + alpha_ep * r_ep
F_value = -r_t # if we view F as a loss
# 4. Compute gradient of F w.r.t. parameters xi
grad_F = compute_gradient_of_F(F_value, xi, R)
# 5. Construct G(xi) metric via Wasserstein geometry
# This involves:
# - Defining a graph of rules with ground metric omega_ij
# - Computing a weighted Laplacian L(p(xi))
# - G(xi) = J_xi^T L(p(xi))^dagger J_xi
G_xi = compute_wasserstein_metric(xi, R)
# 6. Update xi using the Wasserstein natural gradient step
delta_xi = - h * invert(G_xi) * grad_F
xi = xi + delta_xi
# 7. Update rewrite rules R according to new xi distribution
R = update_rules_from_params(xi, R)
# 8. (Optional) Resample environment or input for next iteration
G_in = next_input_state(env)
\end{verbatim}

\paragraph{Details of Key Steps:}

\begin{itemize}
  \item apply\_rules(G\_in, R) : Uses pattern matching to apply rewrite rules from R to the input graph until no more rewrites apply, yielding a predicted output graph.
  \item derive\_distribution(G\_out, R) : Computes a distribution over possible outputs (or internal states) based on how $R$ could have generated G\_out.
  \item KL\_divergence(obs\_dist, pred\_dist) : Standard KL divergence between observed and predicted distributions.
  \item compute\_gradient\_of\_F(F\_value, xi, R) : Numerically approximates or analytically computes how changing xi affects $F$ (requires a parameterization of how xi controls rule selection probabilities).
  \item compute\_wasserstein\_metric(xi, R) : Constructs the ground metric $\omega_{-} \mathrm{ij}$ over rewrite rules, forms Laplacian, computes pseudo-inverse, and then forms G(xi).
  \item update\_rules\_from\_params(xi, R) : Adjusts the set or weights of rewrite rules so that applying them stochastically corresponds to the distribution p(xi).
\end{itemize}

\subsection{Example Scenario: Virtual Bug with Grabber Arm}

Let us illustrate these abstract ideas via a slightly less simplistic example than our above three-cell 1D lattice: a virtual bug in a 2D grid seeking to find food and avoid poison.

\paragraph{Environment Setup:}
\begin{itemize}
  \item A 2D grid: Some cells contain food (F), others are empty (N), and some have poison (P).
  \item The bug sees a local input pattern: its own position $(x, y)$, what's in the cell it faces, and what's in its hand (empty or holding something).
  \item Actions: Move forward, move backward, turn left, turn right, grab an item if available.
\end{itemize}

\paragraph{Initial Rewrite Rules:}
\begin{itemize}
  \item Initially, rewrite rules may be extremely generic, e.g.:
  \item $r_{1}$ : If state pattern includes (Bug sees cell C), produce action (Move Forward).
  \item $r_{2}$ : If state pattern includes (Bug sees cell C), produce action (Turn Left). These rules don't differentiate between F, N, or P. The distribution over these rules might be uniform, causing random actions.
\end{itemize}

\paragraph{Adapting via Discrete ActPC:}
\begin{itemize}
  \item After several episodes, the bug receives different outcomes:
  \item When it moves onto a cell with food and grabs it, eventually obtaining a positive env\_reward.
  \item When it grabs poison, it suffers a negative env\_reward.
  \item When it does random moves, it sees unpredictability and higher error.
\end{itemize}

Through the iterative process:

\begin{enumerate}
  \item The predicted and observed distributions $p_{t}$ and $q_{t}$ diverge if the bug's actions are not aligned with acquiring food.
  \item The system computes $e_{t}$ and thus receives negative instrumental reward when predictions fail and positive when it finds a stable action sequence leading to food.
  \item Applying Wasserstein updates: The parameter $\xi$ shifts probability mass towards rewrite rules that were activated during successful outcomes. For example, if from a certain state (Bug sees F ahead), choosing (Move Forward, Grab) leads to high reward and low surprise, the update will shift mass to a specialized rewrite rule:

$$
r_{\text {food }}:(\text { Bugsees } F) \rightarrow(\text { MoveForward, Grab })
$$
\end{enumerate}

By defining the ground metric $\omega_{i j}$, similar rules (like those involving small modifications of Move Forward or conditions on F) are easier to reach from the initial rules, guiding a smoother transition in rule space.

Over time, the rewrite rules become more specialized:

\begin{itemize}
  \item From (Bug sees F), it learns to (Move Forward, Grab).
  \item From (Bug sees P), it may learn to (Turn Right) or (Move Away).
\end{itemize}

As the distribution over rules evolves, the agent's policy improves, leading to consistent food gathering and minimized error. The discrete Laplacian ensures that changes in rule distributions account for the structural relationships encoded by $\omega_{i j}$.

\subsection{A Slightly More Complex Toy Example}
In the above example, the system adapted by becoming more specialized -- e.g., learning a rule like "If there is an item at ( $x, y$ ) that looks exactly like known food, then go and grab it."  Let us now consider a similar but slightly subtler scenario: food and poison share common low-level features (color, shape, texture) and cannot be distinguished by any single feature alone. Instead, it's a combination of features that determines whether something is food or poison. The system must learn to classify items by observing which combinations lead to rewards (food) and which lead to penalties (poison).

In essence, the rewrite rules must evolve to perform something akin to a logical classification function over feature combinations, rather than merely refine single-step mappings. This involves introducing abstraction: the network should create rules that identify and leverage combinations of features, and perhaps form intermediate symbolic categories (e.g. "FeaturePatternA") that can then be used in higher-level rules.

\subsubsection{ Initial State (Simple Rules):}

Initially, the system might have only simplistic rewrite rules like:

\begin{itemize}
  \item r\_generic: (Sees item at (x,y), features: \{F1, F2, F3...\}) -> (Move Forward, Grab) or
  \item r\_generic\_poison: (Sees item at (x,y), features: \{F1, F2, F3...\}) -> (Move Backward)
\end{itemize}

These rules are too generic. They do not differentiate the subtle combinations of features. The network tries these rules but obtains inconsistent results. Sometimes it grabs poison and is penalized, sometimes it grabs food and is rewarded. The resulting prediction error and reward mismatch drive rule adaptation.\\

\subsubsection{ Local Rewriting and Abstraction Emergence}

 The system explores local modifications to rules. In addition to specializing rules (e.g., restricting a rule to apply only if a certain feature is present), the system can also compose and abstract features. For instance, from an initially flat set of features \{Color=Red, Shape=Round, Texture=Smooth\}, it might introduce new intermediate labels in its rule conditions:

Example evolutionary steps:

\paragraph{Start to form conditional patterns:}

\begin{verbatim}
r_new1: (Sees item with (Color=Red AND Shape=Round)) -> 
define intermediate label PatternA
\end{verbatim}

\noindent This creates a new node/sub-metagraph pattern representing the combined concept "PatternA" that stands for the conjunction of (Red, Round). Similarly, another rewrite step might produce:

\begin{verbatim} 
r_new2: (Sees item with (Color=Green AND Shape=Square)) ->
 define intermediate label PatternB
\end{verbatim}

\noindent Over time, these intermediate labels become "category" nodes inside the metagraph. They represent learned abstractions-feature combinations that consistently correlate with certain outcomes.

\subsubsection{Hierarchical Rule Sets} 

The network can then rewrite high-level rules in terms of these newly created intermediate patterns:

\begin{verbatim}
r_food_candidate: (Sees item with PatternA AND Texture=Smooth) 
-> classify as FOOD_TYPE_A

 r\_poison\_candidate: (Sees item with PatternB AND Texture=Rough) 
 -> classify as POISON\_TYPE\_B
\end{verbatim}

\noindent The idea is that the rewrite rules start encoding conditional checks that go beyond direct observation. They build up a small hierarchy of category nodes that represent combinations of features. In turn, these categories form a bridge between raw features and the final "food or poison" decision.\\

\subsubsection{ Reward-Driven Abstraction Refinement}

During RL-like learning, the system tries actions based on these categories:

\begin{itemize}
  \item If (PatternA, Texture=Smooth) => FOOD\_TYPE\_A consistently leads to gaining reward when the system takes a "grab" action, the rule that classifies (PatternA, Texture=Smooth) as FOOD becomes more probable and stable.
  \item If (PatternB, Texture=Rough) leads to penalties when grabbed, the system will rewrite or adjust the rules so that items matching that combination are treated as poison.
\end{itemize}

\noindent Over many episodes, the rewrite rules evolve from a flat, feature-level decision-making approach to a more layered, hierarchical approach:

\begin{itemize}
  \item Bottom layer: Rules that combine raw features into intermediate patterns.
  \item Middle layer: Rules that combine these patterns with additional features to form stable category labels (e.g. FOOD\_TYPE\_A, POISON\_TYPE\_B).
  \item Top layer: Rules that map these category labels to actions (grab, avoid).
\end{itemize}

\subsubsection{ Mechanics of Abstraction via Rewriting}

The rewriting system can have meta-rules that govern how new categories are introduced. For example:

\begin{itemize}
  \item If a certain conjunction of features frequently leads to a higher reward, the system proposes a new category node to represent that conjunction.
  \item If a certain feature combination leads to unpredictable outcomes, it tries different factorization patterns (splitting the combination into sub-patterns, removing or adding features) and evaluates which factorization yields lower overall prediction error.
\end{itemize}

\noindent This process could be guided by the same information-theoretic and reward-based signals that drive basic rewrite selection. Instead of only refining existing rules, the system now also rewrites the form of the conditions by grouping features, effectively performing a kind of discrete "feature extraction."

\subsubsection{ Abstract vs. Specialized}

Initially, specialization looks like the easiest path: restrict a rule to a single known feature. However, specialization alone can't solve the classification problem if no single feature is discriminative. The rewrite system must "invent" or "discover" intermediate abstractions by combining features. This is akin to feature engineering at the symbolic level.

When a single rule specialized to "Red items" still yields inconsistent rewards, the system tries a local rewrite step that says: "Consider both Color and Shape together." Another rewrite step might say: "Consider Color, Shape, and Texture." Among these attempts, the system retains those rewrites that reduce KL divergence between predicted and observed distributions and/or improve the expected reward.

Through this iterative process, what emerges is a set of more abstract rules: they no longer say "If Color=Red, then grab." Instead, they say "If Color=Red AND Shape=Round (PatternA), and also Texture=Smooth, then treat this as a type of food." Such a rule effectively classifies the item as food by using a combination of features rather than a single specialized condition.\\

\paragraph{Scaling Up} In a more scalable system, these abstractions might become layered, with rewrite rules forming a hierarchy: lower-level rewrite rules define intermediate feature patterns, mid-level rules define categories, and top-level rules map categories to actions. The search over rewrite rules (via local modifications tested for their impact on prediction error and reward) leads to a self-organized taxonomy of item types.

\subsubsection{Conclusion:}

By allowing rewrite rules not only to specialize but also to abstract  -- i.e., to form new intermediate symbolic categories representing feature combinations -- the discrete ActPC-Chem framework  can effectively learn classification functions. Initially random or too-generic rules evolve into a structured set of conditional rewrites that classify food vs. poison items based on the learned distribution of features and outcomes. This process leverages the same fundamental principle of local, error-driven rewriting, but extends it to discovering combinational patterns that yield meaningful conceptual categories.

\section{Integrating Discrete And Continuous Predictive Coding for Neural-Symbolic Robotics}

Now we explore how this novel ActPC variant might work together with traditional continuous ActPC in cases involving a mix of cognition, perception and action.    For concreteness we consider a particular hybrid setting, extending the virtual bug example given above: A real (robotic) bug with a camera (vision) and a motorized grabber arm. 

In the approach we consider, high-level planning and strategy reside in the discrete rewrite-rule network, while two continuous predictive-coding neural networks handle:

\begin{itemize}
  \item Visual Perception Network: Understanding the environment's visual patterns and mapping raw pixels to symbolic features.
  \item Arm Dynamics Network: Handling continuous joint angles, torques, and tactile feedback, translating the discrete actions (e.g. "move arm toward object") into precise low-level motor commands.
\end{itemize}

This hierarchical approach, where the discrete network sets plans and the continuous networks implement them, can exploit the strengths of both discrete and continuous predictive-coding methods.

\subsection{Conceptual Framework}

In the ActPC-Chem approach pursued here, the discrete network is a metagraph $G$ representing states and actions as patterns, with rewrite rules $R$ that map input patterns to output patterns. This will serve as a high-level planning and abstraction layer for our robotic bug.  It will work together with two Continuous Predictive-Coding Networks:

\begin{enumerate}
  \item Vision Network: A hierarchical predictive coding model (e.g., following Ororbia's NGC/ActPC methods) that processes camera images and reduces them to a symbolic representation (like "object at location X is food").
  \item Arm Dynamics Network: Another predictive coding model that receives a desired action (like "grab item at coordinates") and handles the continuous control signals to actuators, learning to predict and correct its own motor output errors.
\end{enumerate}

The discrete network receives processed symbolic inputs from the Vision Network and issues high-level commands for the Arm Dynamics Network, bridging perception and actuation.

\subsection{Rough Formalization}

We may formalize this hybrid architecture roughly as follows.

For the discrete part,

\begin{itemize}
  \item Distribution over rewrite rules $p(\xi)$.
  \item Error measure $e_{t}=D_{K L}\left(q_{t} \| p_{t}\right)$ as before.
  \item Combined reward $r_{t}$ and gradient update using Wasserstein natural gradient:
\end{itemize}

$$
\xi_{k+1}=\xi_{k}-h G\left(\xi_{k}\right)^{-1} \nabla_{\xi} F\left(p\left(\xi_{k}\right)\right)
$$

For the continuous part,

\begin{itemize}
  \item For the Vision Network, let $z$ represent neural state variables predicting image features, and let $e$ represent prediction errors. Neural states update via:

$$
z \leftarrow z+\beta(-\gamma z+(E \cdot e) \otimes \partial \phi(z)-e)
$$

\noindent which is analogous to Ororbia's neural generative coding updates.

  \item Similarly, for the Arm Dynamics Network, continuous predictive coding handles joint angles, forces, and tactile feedback, iteratively minimizing motor prediction errors and refining low-level control signals.
\end{itemize}

To integrate these components,

\begin{itemize}
  \item The discrete network outputs symbolic commands such as "Move toward cell ( $x, y$ )" or "Attempt to grab object."
  \item The Vision Network transforms camera input into a symbolic state that the discrete network uses as $G_{t}^{i n}$.
  \item The Arm Dynamics Network takes the discrete network's chosen action and generates low-level continuous motor commands. Prediction error in the arm network ensures stable and precise arm movements.
\end{itemize}

\subsection{Crude Pseudocode Sketch}

A crude procedural sketch of code for doing initial experimentation with approach might be something like:

\paragraph{Main Loop:}
\begin{verbatim}
loop:
    # Vision processing
    raw_image = robot_camera_capture()
    vision_state = Vision_Net_infer(raw_image) # continuous PC inference producing
symbolic representation
    G_in = construct_input_subgraph(vision_state)
    # Discrete inference
    predicted_output = apply_rules(G_in, R)
    predicted_dist = derive_distribution(predicted_output, R)
\end{verbatim}

\begin{verbatim}
    # Execute action via Arm_Net
    # Suppose predicted_output -> symbolic action: (Move_to x,y; Grab)
    motor_cmd = Arm_Net_compute_actions(predicted_output) # continuous PC-based
motor control
    execute_motor_cmd(motor_cmd)
    # Observe environment outcome
    observed_dist = observe_env_distribution(robot_sensors)
    e_t = KL_divergence(observed_dist, predicted_dist)
    env_reward = compute_env_reward(robot_sensors) # +1 for grabbing food, -1 for
poison, etc.
    r_int = -e_t
    r_ep = compute_surprise(observed_dist, predicted_dist)
    r_t = alpha_int * r_int + alpha_ep * r_ep + env_reward
    F_value = -r_t
    grad_F = compute_gradient_xi(F_value, xi, R)
    L_p = construct_measure_dependent_laplacian(R, p(xi), omega_ij)
    G_xi = compute_G_xi(L_p, J_xi)
    delta_xi = -h * invert(G_xi) * grad_F
    xi = xi + delta_xi
    R = update_rules_from_xi(xi, R)
\end{verbatim}

In parallel, Vision\_Net and Arm\_Net continuously run their own predictive coding updates:

\begin{verbatim}
Vision_Net_update() # Minimizing visual prediction errors
Arm_Net_update() # Minimizing motor and tactile prediction errors
\end{verbatim}

These run at each timestep or sensory update cycle.

Of course, this sort of test implementation would be much more simplistic and less flexible than an AGI-oriented deployment of these same mechanisms within an overall cognitive architecture such as PRIMUS.

\subsection{Physical Robot Bug Example}

Consider a tabletop robot with a camera for "eyes" and a small grabber arm. On the table, we have:

\begin{itemize}
  \item Food Items (F): Robot should pick these up for reward.
  \item Neutral Items (N): Picking these yields no reward.
  \item Poison Items (P): Picking these yields a negative reward.
\end{itemize}

The action, perception and learning process involved in our hybrid ActPC approach then looks something like:

\paragraph{ Low-Level Perception (Vision\_Net):}. The camera image is complex and continuous. The Vision\_Net, a predictive coding neural network, transforms this high-dimensional input into a simplified representation like:

\begin{verbatim}
vision_state = { (type=F, x=2, y=3), (type=P, x=5, y=1) }
\end{verbatim}

The network predicts feature patterns at multiple hierarchical levels, adjusting its internal states and connections to minimize prediction error.

\paragraph{High-Level Planning (Discrete Rules):}. Given vision\_state, the discrete network matches it against rewrite rules:

\begin{verbatim}
r_food: if sees food at (x,y), propose action (move_toward(x,y), grab)
r_poison: if sees poison, propose (move_away)
\end{verbatim}

Early on, these rules are crude. Over trials, the Wasserstein-guided updates shift probability mass to rules that lead to successful food grabs and away from rules that cause interaction with poison.

\paragraph{Low-Level Control (Arm\_Net):}. When the discrete rule says "move\_toward(2,3), grab," the Arm\_Net translates this into continuous joint angle adjustments. Predictive coding in Arm\_Net ensures smooth, stable movements, dynamically correcting errors if the arm overshoots or encounters unexpected resistance.

\paragraph{Feedback and Adaptation:}. After executing actions, the robot observes outcomes. If it successfully grabs food, it reduces error and gains reward. This positive outcome biases the discrete distribution of rules to favor r\_food -like transformations. The Vision\_Net improves its predictions of item positions, and the Arm\_Net refines motor control for more precise grabbing, all through their respective prediction-error minimization loops.

\subsection{Propagating Prediction Errors Across Discrete and Continuous Subnetworks}

To briefly recap: We have introduced a discrete ActPC framework that updates rewrite-rule distributions using a Wasserstein natural gradient on a measure-dependent Laplacian. We then extended the approach to a hybrid scenario: a robot bug that uses two continuous predictive-coding networks to handle complex perception and motor control tasks. The discrete network focuses on higher-level planning and strategy (which items to approach and grab), while continuous networks manage the fine-grained details of vision and arm dynamics. This layered approach leverages the strengths of both discrete, symbolic reasoning and continuous, predictive-coding-based adaptation, potentially scaling to more complex real-world tasks.

By integrating the discrete approach with Ororbia-style continuous predictive coding networks, we illustrate a plausible architecture for hierarchical control. The discrete rewrite-rule planner selects strategic actions, while the continuous networks handle domain-specific complexities. This toy example suggests a pathway to scalable, hybrid solutions mixing discrete program-like policies with biologically inspired neural predictive coding.

Within this framework, one becomes curious about the specifics of how the use of PC (probabilistic error tracking and correction etc.) in the continuous perceptual and action networks might work together with the use of PC in the discrete network.   The particulars will of course depend on how all this is actually implemented, but one can tell an interesting "just so story" at the thought-experiment level.
 
With this in mind, we now give a detailed, step-by-step explanation of how  error tracking and correction might flow through  the three-layered system -- composed of a  discrete ActPC-inspired network of  rewrite rules,  and two continuous predictive coding (PC) 
networks for perception and motor control -- in the example physical robot bug scenario.

Our detailed setup is as follows:

\paragraph{Vision Network (Continuous PC):}
\begin{itemize}
  \item Input: Camera images of the environment, containing food, neutral items, and poison items.
  \item Output: A perceptual representation summarizing salient features (e.g., positions and types of items).
  \item Method: Predictive coding continuously adjusts internal neural states (and possibly synaptic weights) to minimize the difference between predicted sensory features and actual sensory input.
\end{itemize}

\paragraph{Discrete Rewrite-Rule Network (Discrete ActPC):}
\begin{itemize}
  \item Input: Symbolic representation from the Vision Network (e.g., "food at (x,y)", "poison at (u,v)").
  \item Output: High-level action plans (e.g., "move to (x,y), then grab").
  \item Method: Maintains a probability distribution over rewrite rules. Prediction error is measured in terms of information-theoretic divergence (e.g., KL divergence) between predicted outcomes and observed outcomes. Rewards modulate the selection and evolution of rewrite rules.
\end{itemize}

\paragraph{Arm Dynamics Network (Continuous PC):}
\begin{itemize}
  \item Input: A desired high-level action from the discrete network (e.g., "grab at (x,y)").
  \item Output: Continuous joint commands and torque signals for the robot's actuators.
  \item Method: Predictive coding aligns predicted arm states (joint angles, tactile feedback) with actual feedback. It continuously corrects motor outputs to reduce error.
\end{itemize}

Given this setup, the flow of Error Propagation and Correction looks something like:

\paragraph{ Initial Perceptual Processing (Vision Net):}
The camera provides raw pixel data. The Vision Network has an internal generative model that predicts what it should see next, given its current state and top-down predictions. This model might say, "Given the bug's position and known configuration of the environment, I expect to see a food item at a certain location in the image."

\paragraph{ Error Signal in Vision Net:}
The Vision Network compares predicted sensory features (generated internally) with the actual image input. Any mismatch is a visual prediction error signal. For instance, if the network predicted food at $(x=2, y=3)$ but the camera shows no item there, the network registers a prediction error.

\paragraph{Correction in Vision Net:} The Vision Network adjusts its internal states (neural activations representing latent variables) and possibly its weights to reduce this error. Over time, it forms a more accurate representation of what is actually present in the environment.

Once stabilized, the Vision Network outputs a symbolic representation—for example, it might confidently report: "Food item at $(2,3)$." This representation is now less errorful and more trustworthy after multiple inference cycles have minimized internal prediction error.

\paragraph{High-Level Planning (Discrete Network):}
The discrete rewrite-rule network takes the symbolic input from the Vision Net: "Food at $(2,3)$ " and possibly "Poison at $(5,1) . "$ It attempts to predict the outcome of certain action choices. For instance:

\begin{itemize}
  \item If the network currently prefers a rewrite rule that says: (See F at $(2,3)$ ) -> (Move to $(2,3)$; Grab), it predicts that following this rule will lead to obtaining food and a reward.
  \item Error Signal in Discrete Network: The discrete network forms predictions about the next observed state (e.g., after moving and grabbing, it expects to confirm that the grabbed item is food, yielding positive reward and stable future states). Once the action is taken and the environment updates, the network observes actual outcomes. If the expected positive outcome doesn't occur—maybe the bug ended up at the wrong cell or grabbed nothing-there is a symbolic-level prediction error: the observed result differs from the predicted result distribution.
\end{itemize}

This error is measured as an information-theoretic divergence between what the rules predicted and what was observed. Additionally, if no reward is obtained when expected, or if a penalty is received, the discrepancy is also noted.

\paragraph{Correction in Discrete Network: }The network uses the Wasserstein natural gradient and local rule rewrites to adjust the distribution over rewrite rules. If a rule led to high error or low reward, the probability mass shifts away from that rule's configuration. If another rule (e.g., a slight modification of the original one) would better align predictions with observed outcomes (and yield food reward), the distribution moves towards it.

Over many trials, the discrete network "discovers" rewrite rules that more reliably yield correct and rewarding outcomes-effectively minimizing higher-level symbolic and reward-related prediction errors.\\

\paragraph{Motor Control (Arm Dynamics Net):} Once the discrete network selects a plan—say "move the arm toward $(2,3)$ and grab"-the Arm Dynamics Network receives this as a target. The arm network predicts the necessary joint angles, torques, and resulting sensory feedback (proprioception, tactile sensors) from executing this action.

\begin{itemize}
  \item Error Signal in Arm Net: The arm network has its own generative model predicting what the arm's sensors (joint encoders, tactile sensors in the gripper) should report if it correctly executes the move-grab action. If the arm moves and encounters unexpected resistance, overshoots the position, or fails to detect the item in the gripper, the predicted sensor values do not match the actual sensor readings. This mismatch is the arm-level prediction error.
  \item Correction in Arm Net: The Arm Dynamics Network continuously adjusts the motor commands. If it predicted that a certain torque would position the arm at $(2,3)$ but sees that the arm ended up at $(2.5,3)$, it updates its internal states and control signals to correct the position. This may happen in real-time at a high control frequency, rapidly minimizing low-level prediction errors until the arm stabilizes at the desired coordinates and successfully grabs the item.
\end{itemize}

\paragraph{Interaction Among the Three Layers:}
\begin{itemize}
  \item Bottom-Up Error Transmission: From the environment upward:
  
  \begin{itemize}
  \item The Arm Net sees a difference between intended and actual arm positions and corrects it, reducing the local motor error. This correction ensures that the discrete action chosen (e.g., "grab at $(2,3)$ ") is physically realized.
  \item If the Arm Net still fails to achieve the desired action (perhaps the item was at $(2,3)$ visually, but actually it's slightly off), the discrete network eventually sees that its predicted outcome (food grabbed) did not materialize. This sends an error signal up to the discrete network, causing it to reconsider which rule it should use next time.
  \item If the Vision Net's representation was off to begin with (e.g., it mis-labeled a poison item as food), the discrete network will select actions that don't yield expected reward. Over time, this leads the discrete network to maintain pressure on the Vision Net to improve its accuracy. The Vision Net will, in parallel, refine its internal representations to reduce persistent visual prediction errors.
    \end{itemize}
  \item Top-Down Error Modulation: From the discrete planner downward:
    \begin{itemize}
  \item When the discrete network's chosen action fails, it adjusts its rewrite rules and distribution over them. This can mean that next time, it selects a different target location or a different grabbing strategy.
  \item Different actions chosen by the discrete network give the Arm Net different sensory predictions to fulfill, prompting the Arm Net to learn a richer internal model of its capabilities. Similarly, the Vision Net is influenced by the world changes induced by the discrete network's action choices.
    \end{itemize}
  \item Equilibrium of Error Minimization: Over repeated interactions:
    \begin{itemize}
  \item The Vision Network converges to stable, low-error representations of the environment's layout.
  \item The Discrete Network converges to a set of rewrite rules that reliably produce sequences of actions leading to food and avoiding poison.
  \item The Arm Dynamics Network learns a smooth, stable mapping from high-level intended actions to low-level motor controls, minimizing proprioceptive and tactile prediction errors.
    \end{itemize}
\end{itemize}

Thus, all three networks engage in a synergistic cycle of error correction:

\begin{itemize}
  \item The Vision Net reduces sensory prediction error, providing clearer symbolic input.
  \item The Discrete Net reduces symbolic and reward-related prediction error by adjusting its rule set.
  \item The Arm Net reduces motor execution errors to ensure actions match intended plans.
\end{itemize}

Over time, this layered error-correction feedback loop leads to coherent, adaptive behavior: the robot bug learns to perceive objects accurately, choose beneficial actions strategically, and execute those actions precisely.

\section{Integrating Symbolic AI Into ActPC-Based Algorithmic Chemistry}

One advantage that accrues from doing ActPC in a discrete setting, and in particular in a rewrite-rule context, is that this makes it very natural to integrate ActPC with various symbolic-AI mechanisms for reasoning, learning, concept creation and so forth.

This is a large and diverse topic, and we will explore it here only to a limited extent, looking at the examples of the AIRIS causal rule learning algorithm and the PLN uncertain logical inference approach.  From these examples it will be reasonably clear how to integrate other PRIMUS symbolic AI algorithms like concept blending and evolutionary program learning, according to a broadly similar pattern.

\subsection{Integrating AIRIS}

AIRIS (Autonomous Intelligent Reinforcement Inferred Symbolism) \cite{cook2024autonomous} provides a novel algorithm for building causal reasoning in autonomous agents. It learns causal rules from the environment dynamically during interaction, enabling flexible, transparent, and data-efficient learning compared to traditional reinforcement learning (RL). 

Key features of AIRIS include:

\begin{itemize}
\item Rule-Based Learning: Learns expert system-like rules dynamically from environment observations.
\item Causal Reasoning: Builds a graph of causal states (State Graph) for planning and action.
\item Adaptability: Can handle changing objectives or unseen situations by updating its learned rules.
\item Transparency: Provides a scrutable world model for debugging and user control.
\end{itemize}

We will give here a speculative but moderately detailed conceptual and algorithmic description of how to form an AIRIS-ActPC hybrid using the discrete rewrite-rule framework. This approach integrates AIRIS's causal rule inference and state graph construction with ActPC-Chem's predictive-coding-inspired updates and optimal transport-based gradient steps. Both symbolic (AIRIS-like) and predictive coding (ActPC-like) aspects operate through rewrite rules in the same metagraph, allowing for a unified representation and learning mechanism.

\subsubsection{Conceptual Integration}

\paragraph{Unified Metagraph Representation:}. We maintain a single metagraph $G$ whose nodes represent states or concepts and whose edges represent rewrite rules. Each rewrite rule transforms an input pattern into an output pattern. In this hybrid system:

\paragraph{AIRIS-Style Rules (Causal/Symbolic):}
These rules capture causal relations: "If conditions $\{C\}$ hold, then after action $a$, result $S$." They represent learned symbolic knowledge like "If the agent is at (x,y) and takes action 'move north', it ends up at ( $x, y+1$ ) with certain confidence."

\paragraph{ActPC-Style Rules (Predictive/Distributions):}
These rules model predictions and uncertainties. Rather than stating a single deterministic outcome, they define a probability distribution over possible next states. They rely on minimizing prediction error, balancing epistemic and instrumental signals, and can adapt their structure and probabilities using the optimal transport-based Wasserstein gradients.

In practice, each rewrite rule in the metagraph may have a structure like, say

$r_{i}:$ (Input Pattern; Conditions; Action) $\xrightarrow{\text { Probabilities }}$ (Output Pattern)

Some rules (learned from AIRIS-like logic) might be more deterministic and symbolic, while others (ActPC-like) might express distributions over multiple possible outcomes. Over time, all rules evolve to better represent both the causal structure (AIRIS) and the predictive aspects (ActPC).

\paragraph{Single Space for Both Symbolic and Predictive Layers:}.  Instead of having separate symbolic and neural modules, here everything is encoded as rewrite rules and patterns.  The distinction between AIRIS and ActPC is conceptual:

\begin{itemize}
  \item AIRIS-like Functionality: Introduce new rules or adjust existing rules when a discrepancy between expected and observed outcomes is discovered. These rules become more causal and explain sudden changes or new conditions.
  \item ActPC-like Functionality: Continually adjust probabilities and conditions of rules to reduce prediction error and incorporate optimal transport geometry for smooth transitions in rule space.
\end{itemize}

\noindent but at the implementation level, there may not always be a strict distinction between the two components.

\paragraph{Optimal Transport on Rule Space:}. The discrete ActPC approach introduced a measure-dependent Laplacian and Wasserstein natural gradients to find smoother rule updates. Now, when integrating AIRIS:

\begin{itemize}
  \item The set of rewrite rules and their probabilities $p(\xi)$ form a discrete probability distribution over "how the agent believes the environment works."
  \item AIRIS's new rule introductions and confidence adjustments (symbolic reasoning) modify the structure and conditions of these rules.
  \item The ActPC Wasserstein step adjusts the distribution over these rules, ensuring that each update respects the underlying cost structure $\omega_{i j}$ in the rule space (where $\omega_{i j}$ measures the 'distance' or 'difference' between rules $r_{i}$ and $r_{j}$ ).
\end{itemize}

\subsubsection{Learning Process}

Given the above setup, the learning process of the combined ActPC-Chem / AIRIS system would look something like

\paragraph{Initial State:}

\begin{itemize}
  \item Metagraph $G$ with a small set of generic rewrite rules, some representing trivial predictions (e.g., "move forward leads to forward motion") and others representing no-op or random actions.
  \item Parameter vector $\xi$ encodes a probability distribution over these rules.
\end{itemize}

\paragraph{ Observation and Action:} At each step:

\begin{itemize}
  \item The agent's current state subgraph $G_{t}^{i n}$ is known.
  \item The rewrite rules $R_{t}$ are applied to predict next states and choose actions. AIRIS-like reasoning tries to find a path (a sequence of rewrites) to a goal state. ActPC-like prediction tries to match predicted and observed distributions.
\end{itemize}

\paragraph{Prediction vs. Observation:}  After executing an action and observing the next state:

\begin{itemize}
  \item Compare predicted distribution $p_{t}(m)$ of outcomes with actual outcome $q_{t}(m)$.
  \item If a discrepancy arises, the system:
  \begin{itemize}
  \item Uses AIRIS logic to create or update a rule to explain the unexpected change. For example, if the agent thought "move north" would result in state A but got state $B$, a new rule that refines conditions or outcomes is introduced.
  \item Uses ActPC logic to adjust the probabilities and parameters of rules, minimizing KL divergence (or other information-theoretic errors) and integrating rewards.
\end{itemize}
\end{itemize}

\paragraph{Causal Inference (AIRIS) Within the Metagraph:}

AIRIS normally forms a state graph and learns rules IF \{C\} THEN S. Here, these rules are just special rewrite rules in the metagraph:

$$
r_{AIRIS}: IF\{C\} AND ACTION=a \Longrightarrow NEXT_{-} STATE=S
$$

\noindent I.e.,

\begin{itemize}
  \item When a discrepancy is found, a new rule is introduced:
  \item Conditions $\{C\}$ might be detected by analyzing which aspects of the previous state differ from the observed next state.
  \item Statements $S$ encode the newly discovered result of applying action $a$ under conditions $\{C\}$.
  \item Confidence in these rules is updated by counting how often conditions match and outcomes come true.
\end{itemize}

\paragraph{Predictive Coding (ActPC) on the Same Rules:}

 Each rewrite rule can have associated probability or confidence weights.  Prediction error looks l ike

$$
e_{t}=D_{K L}\left(q_{t} \| p_{t}\right)
$$

\noindent and rewards combine epistemic (exploration) and instrumental (goal achievement):

$$
r_{t}=\alpha_{i n t}\left(-e_{t}\right)+\alpha_{e p}(\text { surprise measure })+\text { environmental reward }
$$

Parameter updates for the distribution over rules look like:

$$
\xi_{k+1}=\xi_{k}-h G\left(\xi_{k}\right)^{-1} \nabla_{\xi} F\left(p\left(\xi_{k}\right)\right)
$$

\noindent where $G(\xi)$ is computed from the measure-dependent Laplacian on the rule graph and $F=-r_{t}$.

This latter step moves probability mass toward rules that better explain observations and lead to rewards, while respecting the geometry defined by $\omega_{i j}$.

Crude pseudocode for initial standalone experimentation with this sort of process might look like

\paragraph{Pseudocode}

\begin{verbatim}
Initialize metagraph G with an initial set of rewrite rules R  
Initialize parameters xi for the distribution over these rules
loop:
current_state = observe_environment()
# Predict next state and choose action via rewrite rules:
predicted_outcomes = apply_rewrite_rules(current_state, R, xi)
# Select action from predicted distribution (ActPC) + cause-effect planning (AIRIS)
action = select_action(predicted_outcomes)
# Execute action in environment
next_state = environment_step(action)
# Observe outcome distributions q_t(m)
observed_distribution = derive_distribution(next_state)
predicted_distribution = derive_distribution_from_rules(current_state, action, R, xi)
# Compute error and reward
e_t = KL_divergence(observed_distribution, predicted_distribution)
env_reward = compute_environmental_reward(next_state)
r_ep = compute_surprise(observed_distribution, predicted_distribution)
r_t = alpha_int * (-e_t) + alpha_ep * r_ep + env_reward
F_value = -r_t
# AIRIS-like rule updates: If discrepancy is found, create or refine a rule
if discrepancy_detected(predicted_distribution, observed_distribution):
     new_rule = create_or_refine_rule(current_state, action, next_state)
      R = add_rule_to_metagraph(R, new_rule)
# Compute gradient wrt xi for ActPC updates
grad_F = compute_gradient_xi(F_value, xi, R)
# Construct measure-dependent Laplacian L(p(xi)) from rule graph
L_p = construct_measure_dependent_laplacian(R, p(xi), omega_ij)
G_xi = compute_G_xi(L_p, J_xi)
# Update xi via Wasserstein natural gradient
xi = xi - h * invert(G_xi) * grad_F
# Update rewrite rules probabilities and conditions based on updated xi
R = update_rule_distribution(R, xi)
# Over time, AIRIS-like rules become more reliable and ActPC-like predictive
# distributions become more accurate and stable.
\end{verbatim}

\subsubsection{How Abstraction Emerges:}

As AIRIS-like logic introduces rules to explain unexpected transitions, these rules can incorporate more complex conditions. For instance, they can join multiple features into a condition that reliably predicts an outcome. This is how the symbolic, causal layer emerges.  That is,

\begin{itemize}
  \item The ActPC component ensures that even with new rules added, the overall distribution over rules converges to a set that minimizes prediction error and achieves goals. If certain introduced rules do not improve predictive accuracy or reward achievement, their probability is reduced and they may eventually be pruned or replaced by betterfitting rules.
  \item Through iterative rewriting, the system can build hierarchical abstractions. Lower-level rules might define intermediate feature patterns; higher-level rules use these patterns as conditions for actions leading to specific outcomes, merging the symbolic clarity of AIRIS with the adaptive predictive tuning of ActPC.
\end{itemize}

\subsubsection{Recap of ActPC-Chem + AIRIS}

Summing up, in this hybrid AIRIS-ActPC-Chem approach:

\begin{itemize}
  \item Both symbolic causal reasoning and predictive modeling occur through the same rewrite-rule mechanism in a single metagraph.
  \item AIRIS-like operations handle causal discovery and rule creation when discrepancies appear.
  \item ActPC-like gradient-free optimization and optimal transport geometry guide probability shifts among rules, ensuring stable convergence and balanced exploration-exploitation.
\end{itemize}

By unifying symbolic and predictive coding updates in one discrete rewriting framework, this hybrid model leverages the strengths of both AIRIS (causal reasoning, transparency) and ActPC (adaptive, gradient-free learning) to create a powerful, interpretable, and robust learning system.

{\it (Or that's the theory, at any rate...) }

\subsection{ActPC-Chem with AIRIS: A Virtual Bug Example}

To make this a little more concrete, we now give a moderately detailed hypothetical example integrating AIRIS-like causal rule inference and ActPC-like predictive coding in the discrete rewrite-rule framework, applied to the virtual bug scenario. We'll show how this hybrid approach can help the bug learn complex, time-delayed and context-dependent distinctions between food and poison items.

\subsubsection{Scenario Setup}

\paragraph{The Environment:}

\begin{itemize}
  \item A virtual bug moves in a grid world collecting items.
  \item Items appear with various visual/olfactory features (color, shape, smell, etc.).
  \item Some items are always good food.
  \item Some items are poison, but their effects vary:
  \item Delayed Poison: Some items only cause sickness hours (or many time-steps) after ingestion.
  \item Conditional Poison: Some items are beneficial if the bug keeps moving after eating them (e.g., "running digestion" turns them into a good nutrient source), but harmful if the bug sits still after ingestion.
\end{itemize}

\paragraph{The Challenge:}
\begin{itemize}
  \item The bug cannot distinguish which items are safe or dangerous by immediate reward alone.
  \item The bug must discover causal rules that connect item features, actions after ingestion, and temporal delays, to the eventual outcome (health improvement, neutral effect, or sickness).
  \item This requires building complex causal models-something AIRIS is good at—while also dealing with uncertainty and continuously adapting predictions-something ActPC excels at.
\end{itemize}

\paragraph{State Representation using Metagraph with Rewrite Rules:}

In the most straightforward approach, we could say

\begin{itemize}
  \item Nodes represent states or concepts: these might include the bug's position, the items it holds, its current health, how long since it ate a particular item, and whether it is moving or resting.
  \item Edges (rewrite rules) transform an input pattern (current state + action) into an output pattern (next state), often probabilistically.
\end{itemize}

Initially, the bug has very generic rules:

\begin{verbatim}
r_move: (Bug at (x,y), sees item) + Action=Eat -> (Bug holds item, same features)
r_wait: (Bug holds item, t since ingestion < threshold) + Action=Wait -> (Bug holds
item, t+1)
...
r_generic_poison:
r_generic_food:
\end{verbatim}

These rules do not yet differentiate subtle feature combinations or delayed effects.

\paragraph{Causal Discrepancy Detection:}

Suppose that

\begin{itemize}
  \item The bug tries eating a new item with certain features: \{Color=Red, Shape=Round, Smell=Sweet\}.
  \item Immediately after ingestion, nothing bad happens, so the bug's current rewrite rules predict a neutral or beneficial outcome.
  \item Hours later: The bug finds itself sick without any immediate causal explanation from its current rules. The predicted outcome ("all good") diverges from the observed outcome ("sickness").
\end{itemize}

When the bug detects that a previously predicted neutral state ended up being sickness after time passes, it triggers an AIRIS-like causal inference step:

\begin{enumerate}
  \item Identify conditions that were present at the time of ingestion.
  \item Identify new rules that can explain this delayed change:
\end{enumerate}

\begin{verbatim}
r_new_causal: IF {ate item with (Color=Red,Shape=Round,Smell=Sweet) AND after N
steps} THEN Sickness
\end{verbatim}

\paragraph{Incorporating Time and Behavior:}

 If the bug observes that when it eats this item and then keeps moving (e.g., running around), it does not get sick (perhaps it "burns off" the toxin), AIRIS-like reasoning creates a more conditional rule:

\begin{verbatim}
r_conditional: IF {ate Red-Round-Sweet item, AND after N steps, bug was mostly
active} THEN Good (Nutritious outcome)
ELSE IF {ate Red-Round-Sweet item, AND after N steps, bug was mostly inactive}
THEN Poisonous outcome
\end{verbatim}

These new rules are now added to the metagraph, introducing conditional dependencies and temporal delays. The conditions include not just item features, but also the bug's activity pattern and elapsed time since ingestion.

\paragraph{ActPC-Style Predictive Coding on the Same Rules}
As new causal rules are introduced, the system still faces uncertainty. Not every Red-RoundSweet item behaves identically, and some may have different time scales or require different degrees of activity. The ActPC portion helps here by maintaining and adjusting probabilities of these rules:

\begin{itemize}
  \item Prediction Error Minimization: After introducing r\_new\_causal and r\_conditional rules, the system tries to predict the outcome of ingesting certain items. If the predictions still don't match observations (maybe some Red-Round-Sweet items only become poison after 2 hours, others after 3 hours), ActPC's information-theoretic error signals kick in. The system updates the probability distribution over rules to minimize KL divergence between predicted and observed distributions.
  \item Wasserstein Natural Gradient Updates: As multiple candidate rules compete to explain delayed poison effects, the ActPC framework uses the measure-dependent Laplacian and optimal transport metric to shift probability mass smoothly:
  \item If r\_conditional explaining "active running prevents sickness" consistently aligns with outcomes, its probability increases.
  \item If a simpler rule that doesn't consider activity patterns fails to predict outcomes well, probability mass gradually shifts away from it.
\end{itemize}

The combination ensures that, over time, the system does not just add more and more rules -it also filters, refines, and balances them according to predictive accuracy and reward signals.

\paragraph{Temporal and Conditional Complexity: Delayed Effects:}

When poison takes effect hours later, a purely reactive approach (like immediate reinforcement signals) would struggle. The hybrid AIRIS-ActPC system can then use AIRIS-like inference to introduce temporal conditions into rules. For instance:

\begin{verbatim}
r_delayed_poison: IF {ate item with pattern A, and t > T_threshold} THEN
become sick
\end{verbatim}

The use ActPC to manage uncertainty here is clear: eg. perhaps it's not always t > T\_threshold, maybe it's t > T\_threshold+1. Over multiple episodes, ActPC tries variants of these rules (rewrite steps that adjust the temporal threshold) and keeps those that minimize prediction error and maximize reward.

\paragraph{Contextual Conditions:}

 The "running vs. sitting still" condition requires rules that incorporate the bug's own behavior patterns. Initially, the system might have a rule:

\begin{verbatim}
r_guess: IF {ate pattern A item} THEN after T steps become sick
\end{verbatim}

Observing that when the bug moves continuously after ingestion, sickness doesn't occur, triggers AIRIS to refine this rule into:

\begin{verbatim}
r_refined: IF {ate pattern A item, AND inactivity_level > X} THEN after T steps
become sick
ELSE IF {ate pattern A item, AND inactivity_level <= X} THEN no sickness
\end{verbatim}

ActPC ensures that if this refined rule reduces long-term prediction error and leads to better policies (e.g., the bug learns to keep moving after eating ambiguous items to avoid sickness), it becomes more probable and stable in the system.

\paragraph{Iterative Refinement}
\begin{enumerate}
  \item Initial Trials: The bug eats unknown items, half the time it becomes sick after a delay, half the time not. The system is initially confused.
  \item AIRIS Rules Emerge: Observing patterns, AIRIS introduces rules with conditions on item features, time delays, and the bug's activity. For example, a rule emerges like:
\end{enumerate}

\begin{verbatim}
r_complex: IF {Item matches features: Red-Round-Sweet, Ate at time t0}
    AND {At time t0+N, bug's motion pattern = "mostly still"}
    THEN Sickness at t0+N
\end{verbatim}

Another variant:

\begin{verbatim}
r_alternative: IF {Item matches features: Red-Round-Sweet, Ate at time t0}
AND {At time t0+N, bug's motion pattern = "mostly moving"}
THEN Good Nutrition at t0+N
\end{verbatim}

ActPC Probability Adjustment could look like: 

\begin{itemize}
\item Initially, maybe multiple candidate rules are proposed, differing in the exact threshold N or the required inactivity level. ActPC tries each rule out (due to exploratory updates and stochastic rule application). Over many episodes:
\item Afterwards,

\begin{itemize}
  \item Rules that fail to predict outcomes accurately are assigned lower probability.
  \item Rules that consistently match observed transitions and lead to rewards (the bug learns to use them to avoid sickness and gain nutrition) are assigned higher probability.
\end{itemize}

\end{itemize}

The measure-dependent Laplacian ensures that modifications to rules are not random jumps, but structured steps respecting rule-space geometry defined by similarity metrics $\omega_{i j}$.\\

\paragraph{End Result:}. Eventually, the system stabilizes on a set of rewrite rules that capture the causal complexity:

\begin{itemize}
  \item Certain feature combinations + inactivity lead to delayed poisoning.
  \item The same items under different activity conditions yield nutrition.
\end{itemize}

The bug's learned policy, guided by these rules, might be: "If I pick up a Red-RoundSweet item, I must keep moving for the next hour to avoid sickness. If I stop moving, I get sick." This is exactly the kind of subtle causal pattern that AIRIS's rule inference captures, while ActPC refines and stabilizes the distribution of rules to ensure accurate predictions and good performance.

\subsubsection{Conclusion}
In this hybrid AIRIS-ActPC discrete rewrite-rule context, the bug learns complex causal relationships involving delayed and conditional poisoning:

\begin{itemize}
  \item AIRIS contributes by introducing and refining rules that incorporate temporal delays and conditional logic (e.g., activity-dependent digestion).
  \item ActPC ensures these rules are integrated into a predictive coding framework that continuously reduces error and balances exploration (epistemic reward) with goal achievement (instrumental reward).
  \item Together, they enable the bug to discover that certain items are "conditionally good" or "delayed poison," adapting its long-term behavior to these complex causal patterns.
\end{itemize}

This synergy allows for emergent causal inference and stable, accurate predictive modeling in a challenging environment where cause-and-effect relationships span both time and contextual conditions.

\subsection{Integrating PLN Inference}

As an additional hypothetical exploration regarding integration of advanced symbolic cognitive methods into the ActPC-Chem framework, we now flesh out a moderately detailed scenario illustrating how a Probabilistic Logic Networks (PLN) inference engine \cite{goertzel2008PLN} could be integrated into the hybrid AIRIS-ActPC discrete rewrite-rule framework.

Continuing with the "virtual bug" example for simplicity and concreteness, we focus on how PLN's uncertain logical inference capabilities can propose probabilistic abstractions and hypotheses, which then get tested and refined by the ActPC predictive-coding error signals and integrated into AIRIS's causal rule graph. The synergy of these three components (PLN, AIRIS, ActPC) yields a system capable of not just reacting and adapting, but also logically generalizing from past experience to guide future exploration.

Summing up we here consider a system for virtual bug control involving three key components;

\begin{itemize}
\item AIRIS (Causal Rule Inference): AIRIS focuses on discovering and refining causal relationships in the environment. It detects when observed outcomes diverge from expectations and introduces new rewrite rules or conditions that can explain these discrepancies causally.

\item ActPC (Predictive Coding in Discrete Setting): ActPC uses predictive coding principles to assign probabilities to rewrite rules, minimizing prediction error and balancing exploration and exploitation. It tracks how well the current set of probabilistic rules matches observed outcomes, adjusting the probability distribution over rules using optimal transport-based gradient steps.

\item  PLN (Probabilistic Logic Networks):  PLN can reason inductively, abductively, and deductively over uncertain knowledge, generating probabilistic inferences and abstractions. It can take the raw experiences and previously learned rules and guess new, higher-level rules: for instance, "If items have a certain pattern of features, they are probably beneficial if digested while moving."
\end{itemize}

The envisioned combined workflow then looks roughly like:

\begin{itemize}
  \item Experience \& Observation: The bug interacts with its environment, encountering items with various features, ingesting them, and experiencing delayed or conditional effects (some only poison if the bug stops moving afterward, etc.).
  \item Local Causal Updates (AIRIS): When unexpected outcomes occur, AIRIS updates or adds rewrite rules with more complex conditions.
  \item Global Probabilistic Refinement (ActPC): ActPC monitors the global consistency of predictions and outcomes, adjusting the probabilities assigned to each rewrite rule to minimize predictive error.
  \item Abstraction \& Hypothesis Generation (PLN): Using the knowledge the bug has acquired, PLN creates higher-level probabilistic rules. For example, from instances of "red-round-sweet" items that become nutritious if running and poisonous if still, PLN might generalize that "Any item with a particular set of chemical markers (detected indirectly) behaves similarly." PLN can also propose rules about unseen combinations of features, guessing their likely outcomes based on analogies or inductive reasoning.
\end{itemize}

\subsubsection{Integration of PLN with AIRIS and ActPC}

So let's run through this potential integration in more detail.

Consider e.g. PLN as a Hypothesis Generator: Suppose the bug has encountered several items that share only partial feature overlap with previously known food or poison patterns. The bug's current AIRIS/ActPC rule set explains some patterns but leaves others uncertain, leading to persistent prediction errors.

PLN steps in:

\begin{itemize}
  \item Using logical inference (induction), PLN notices that items with high sugar content are generally beneficial if the bug keeps moving after ingestion.
  \item It also notices that items containing a certain organic compound (inferred from partial features) often become poisonous if the bug remains still.
\end{itemize}

PLN forms a new probabilistic hypothesis:

\begin{verbatim}
If (item has FeatureSet X) and (after ingestion, activity pattern = still), then
Probability(Outcome=Poison) = p1
\end{verbatim}

\begin{verbatim}
If (item has FeatureSet X) and (after ingestion, activity pattern = moving),
then Probability(Outcome=Beneficial) = p2
\end{verbatim}

This hypothesis is expressed as a set of conditional rewrite rules with associated probabilities.

\paragraph{Inserting PLN's Abstractions into the Metagraph:} The metagraph that stores rewrite rules now receives new candidate rules from PLN. For example:

\begin{verbatim}
r_pln_hyp: IF {Item Features = X} AND {t steps after ingestion} THEN:
    - With probability p1: leads to Poison if still
    - With probability p2: leads to Benefit if moving
\end{verbatim}

\noindent These rules might not be immediately reliable. They are hypotheses. AIRIS previously discovered conditions for specific known items, while PLN provides a more abstract, general rule that might explain novel items.\\

\paragraph{Testing PLN-Derived Rules with ActPC:}  ActPC monitors the prediction errors. If the bug tries ingesting a new item that fits FeatureSet $X$ :

\begin{itemize}
  \item The system uses both previously learned AIRIS rules (more specific) and the new PLN-proposed rule (more general and abstract).
  \item The predicted outcome is now a mixture of probabilities from different rules (some specific rules learned earlier, and the new general rule from PLN).
\end{itemize}

After the bug acts (e.g., it stays still after ingesting the item), the actual outcome is observed. Suppose the outcome matches the PLN's predicted pattern (the bug gets mildly sick after 1 hour). This reduces prediction error and validates the PLN rule's usefulness.

ActPC then shifts probability mass toward this new PLN-derived rule, reinforcing it. If instead the rule had failed to predict the correct outcome, ActPC's error signals would lower its probability and prompt PLN or AIRIS to consider alternative inferences.

\paragraph{Refining Causal Structure with AIRIS} If a PLN-generated rule is partially correct but not entirely, AIRIS can refine it by adding more nuanced conditions/

E.g. maybe the PLN said FeatureSet X leads to poison if still, benefit if moving. The bug finds that's only true if temperature is below a certain threshold. AIRIS introduces a condition for temperature:

\begin{verbatim}
r_refined: IF {Item Features = X, Temperature < T, inactivity} THEN Poison
    IF {Item Features = X, Temperature >= T, inactivity} THEN Neutral
\end{verbatim}

With each refinement, the system merges PLN's abstract hypotheses with AIRIS's concrete causal inference to yield more accurate and context-rich rewrite rules.

\paragraph{Continuous Feedback Loop:}
\begin{itemize}
  \item PLN: Generates hypotheses about unseen events or novel feature combinations using logical inference and generalization.
  \item AIRIS: Detects discrepancies and refines rules, inserting context conditions that handle discovered exceptions.
  \item ActPC: Provides the "ground truth" check in terms of predictive accuracy and reward acquisition, using error signals to modulate the probability distribution over the entire rule set. This ensures that both new (PLN-proposed) and old (AIRISdiscovered) rules must stand the test of predictive accuracy and environmental reward.
\end{itemize}

\subsubsection{One More Virtual Bug Thought Experiment}

Consider next the following slightly more complex scenario:

\begin{itemize}
  \item The bug encounters a new class of items with a faint odor it never experienced before: FeatureSet X = \{Odor=FaintSweet, Color=Purple, Shape=Round\}.
  \item No existing AIRIS rule matches this item exactly; the bug tries ingestion.
  \item Without PLN, the system would rely purely on trial-and-error combined with local causal refinement. It might take many trials to guess that these Purple-Round-FaintSweet items behave like Red-Round-Sweet items if digested in a warm environment and while running.
\end{itemize}

With PLN, on the other hand, we can have a dynamic like:

\begin{itemize}
  \item PLN sees a pattern: previously, any item with "Sweet-like" odor and "Round" shape turned out beneficial if the bug stayed active post-ingestion. PLN generalizes: "Likely this new faint-sweet-odor item also benefits from post-ingestion movement."
  \item PLN proposes a rule:

\begin{verbatim}
r_pln_generalization: IF {Odor ~ Sweet, Shape = Round} AND {Maintain movement
post-ingestion} THEN Probably beneficial
\end{verbatim}

  \item The bug tries it out. The ActPC module notes a lowered prediction error if the bug moves after ingesting the new item, confirming PLN's guess.
  \item If the item's effect turns out to depend on temperature as well, AIRIS notices discrepancies when the bug tries the same strategy in cooler conditions. AIRIS refines the rule to:
  
\begin{verbatim}
r_pln_airis_refined: IF {Odor ~ Sweet, Shape=Round, Temp < T, Movement post-
ingestion}
    THEN High chance of benefit
        ELSE if Movement but Temp >= T
    THEN Neutral benefit (not as good)
\end{verbatim}

  \item ActPC integrates this refined rule into the probability distribution, maintaining or increasing its probability as long as prediction error stays low.
\end{itemize}

Over multiple episodes, PLN's broad inductive leaps speed up the discovery of good strategies, while AIRIS hones these strategies into context-specific causal rules, and ActPC ensures that only rules that actually lead to consistent and low-error predictions remain prominent.

\subsubsection{Conclusion}
Overall, it seems that by embedding PLN's uncertain logical inference into the AIRIS-ActPC framework, what we may achieve is:

\begin{itemize}
  \item PLN contributes advanced generalization and probabilistic logical inference capabilities, proposing new rules about unseen events or novel combinations of features.
  \item AIRIS ensures these hypothesized rules are tested and, if necessary, refined into more context-sensitive causal patterns.
  \item ActPC provides a continuous quality control loop, using prediction error and reward to regulate which rules are taken seriously and which are downgraded.
\end{itemize}

In this integrated architecture, for instance, a virtual bug doesn't just react and learn incrementally. It also forms probabilistic abstractions and guesses about new items' behaviors, tests these guesses against reality, and continuously refines them. The synergy allows a richer, more adaptive, and more rapidly converging learning process than any of the components could achieve alone.

This sort of system could also serve as a foundation onto which other PRIMUS cognitive methods could be integrated, e.g. MOSES procedure learning, concept blending, and so forth.  The general advantages conferred by these techniques would be manifested in these simple "bug world" scenarios in fairly concrete ways, which however we will not take time to elaborate here, having chosen AIRIS and PLN as the two symbolic PRIMUS algorithms to discuss in moderate detail right now.

\section{Toward an ActPC and Algorithmic Chemistry Based Transformer-Like Prediction Network}

As a complement to the potential utilization of ActPC-Chem for experiential learning experiments with simple agents in virtual or robotic settings, it may also be interesting to experiment with the methodology for larger-scale "narrow AI" ish applications, such as biomedical or financial data analytics, or emulating some of the functions of current LLMs in a fashion more amenable to improvements overcoming some of these system's well-known short-comings in terms of fact-groundedness, creativity and sustained systematic reasoning.

We will sketch here a high-level conceptual architecture outlining how one might construct a transformer-like system -- akin to modern LLMs in high level structure and function, but intended for improved AI capability --  using the combination of methods described above: discrete ActPC for core learning rather than backprop, AIRIS for causal reasoning to reduce hallucinations, PLN for probabilistic logical abstractions, and integrated continuous predictive-coding (PC) neural networks for multimodal perception.  The goal is to convey the rough structure and information flow rather than the implementation details -- this is definitely a pointer in a promising-looking research direction rather than a highly particular architecture one would expect to code up and have working right out of the box!

Traditional transformers (e.g., the ChatGPT models and their main commercial competitors and open source analogues) rely on large-scale backpropagation through a network of attention and feedforward layers to model sequences.   In the approach roughly sketched here, 

\begin{itemize}
\item we replace backprop with a discrete ActPC (Active Predictive Coding) mechanism that updates a probabilistic distribution over rewrite rules that govern token transformations
\item AIRIS-like causal reasoning and PLN-driven probabilistic logic introduce causal and conceptual coherence
\item Continuous PC neural networks handle sensory (e.g., image, sound) embeddings, integrating multimodal input seamlessly.
\end{itemize}

\subsection{Architectural Components}

\subsubsection{Discrete ActPC Core (Instead of Standard Transformer Layers):}

\paragraph{ Rewrite Rules as a Basis of Computation:} Instead of linear projections and backprop-updated weights, we store a set of probabilistic rewrite rules that transform sequences of tokens/states into predicted next tokens/states. These rules form a metagraph playing the role of the neural layers in a standard transformer.

\paragraph{ Attention-Like Mechanism via Rule Selection and Matching:} Instead of scaled dot-product attention, the system uses a "rule matching and application" process. Given a current sequence state, the system selects from a set of candidate rewrite rules-guided by probabilities maintained and updated by ActPC-to determine which rules best predict the next element of the sequence.

\paragraph{ Prediction Error Minimization (ActPC):} Each step, the model predicts the next token or next set of states. When the actual next input is observed, the difference between prediction and reality drives error signals that update the probability distribution over rules. Optimal transport geometry ensures stable, gradient-free adaptation of the rule distribution.

\subsubsection{AIRIS-Like Causal Reasoning Integration:}

\paragraph{ Causal Graph \& Conditions in Rewrite Rules:} AIRIS principles introduce conditional rewrite rules that capture causal relations between tokens or states. For example, certain linguistic constructs might cause certain future patterns. Over time, the system refines these causal rules to reduce nonsensical correlations (hallucinations).

\paragraph{  Context and Structural Coherence:}  If the sequence represents text, AIRIS-like reasoning can detect when certain narrative or logical structures are violated and attempt to introduce new conditional rules ensuring that future predictions respect previously established causal or narrative constraints.

\subsubsection{PLN (Probabilistic Logic Networks) for Abstraction \& Induction:}

\paragraph{ Abstract Probabilistic Generalizations:} PLN provides a layer of reasoning that can propose new rewrite rules at a higher conceptual level. For instance, if the model frequently encounters a pattern of reasoning steps, PLN might introduce a new abstract rule capturing this pattern and assign it a probability.

\paragraph{Reducing Hallucinations and Improving Consistency:} PLN can help recognize that certain transitions are logically inconsistent, even if statistically plausible. It can propose constraints that the ActPC core tries to satisfy by adjusting probabilities of conflicting rewrite rules.

\paragraph{ Inductive Leap for Unseen Patterns:} When faced with novel sequences, PLN can induce rules that generalize from observed patterns, allowing the model to guess what should come next even without direct training examples, relying on logical and conceptual similarities to known cases.

\subsubsection{Continuous PC Neural Networks for Multimodal Input:}

\paragraph{  Multimodal Embeddings via Continuous PC:} Just as standard GPT-like models use learned embeddings for tokens, here we can integrate continuous predictive-coding neural sub-networks for images, audio, or robot movement states. Each modality's input passes through a PC network that minimizes sensory prediction error and outputs a stable latent representation (symbolic or feature-based).

The process then looks like:

\begin{itemize}
  \item Translation into Rewrite Rule Space: The continuous PC networks output feature codes that can be turned into "pseudo-tokens" or symbolic patterns the discrete ActPC framework can handle. For example, an image frame leads to a set of visual features that the rewrite rules treat like tokens.
  \item Bi-Directional Influence: The discrete ActPC core and PLN/AIRIS modules can feed predictions back to the continuous PC networks, guiding them to refine their latent representations to better match the expected context. This is akin to top-down predictive coding in hierarchical generative models.
\end{itemize}

\subsection{Information Flow (High-Level):}

The flow of information within this proposed architecture then looks like:

\subsubsection{ Input Processing:}

\begin{itemize}
  \item Multimodal input (text tokens, image frames, audio signals, robot sensor states) is received.
  \item Continuous PC networks process images/audio/motor states, outputting stable feature sets. Text tokens enter directly as symbolic states.
\end{itemize}

\subsubsection{Rewrite Rule Prediction (ActPC Core):}
\begin{itemize}
  \item The current sequence state (including the just-processed multimodal features) is matched against existing rewrite rules.
  \item Multiple candidate rules predict possible next tokens (in a textual scenario) or next sensory states.
  \item Probability distribution over these rules is shaped by past successes/failures and updated via prediction error minimization.
\end{itemize}

\subsubsection{Causal \& Logical Refinement (AIRIS + PLN):}
\begin{itemize}
  \item If the model's predictions start to diverge from sensible causal relationships (e.g., narrative consistency or logical coherence), AIRIS proposes new or refined rules that explicitly encode these causal constraints.
  \item PLN simultaneously attempts to infer more abstract patterns. For example, if certain narrative progressions or reasoning steps are repeatedly rewarded (low error), PLN generalizes a higher-level rule that can skip intermediate steps and more directly predict logically consistent outcomes.
\end{itemize}

\subsubsection{Error Feedback Loop:}
\begin{itemize}
  \item Once the environment or dataset provides the actual next token (or the bug's next sensory reading), the difference from the predicted distribution feeds back into the system.
  \item ActPC uses this error to shift probability mass among the rewrite rules.
  \item AIRIS checks if the newly introduced causal conditions reduce unexplained discrepancies.
  \item PLN can further refine abstractions or propose alternative logical constraints if errors persist.
\end{itemize}

\subsubsection{ Iterative Refinement and Convergence:}

\begin{itemize}
  \item Over many sequences, just like a GPT refines its weights via backprop, this system refines its rewrite rules and their probabilities via ActPC updates.
  \item AIRIS continuously shapes the causal graph underlying these rewrite rules, preventing persistent logical inconsistencies.
  \item PLN gradually builds up a library of abstract probabilistic rules that can handle novel contexts smoothly.
\end{itemize}

\subsection{Parallels to Transformer Components}

The key aspects of standard transformer neural net architectures may be reflected within this approach as follows:

\begin{itemize}
  \item Layers and Depth: We can stack multiple "layers" of rewrite rules, where higher layers handle more abstract patterns (thanks to PLN and AIRIS) and lower layers handle more direct pattern completions. Each layer refines predictions from the previous, similar to a transformer's stack.
  \item Attention Mechanism: Instead of computing attention scores via dot products, "attention" emerges from how rewrite rules match patterns in the current input. Probabilistic selection of rules that are contextually appropriate functions similarly to attention selecting which parts of the input to focus on.
  \item Feedforward Sublayers: The "feedforward" transformations could be sets of rewrite rules that transform intermediate representations into predicted next tokens, again chosen probabilistically and refined by ActPC rather than by learned dense layers.
\end{itemize}

Overall, then, one sees how  a transformer-like generative model could be built without backprop-based training, relying instead on a synergy of discrete ActPC rule selection, AIRIS causal refinement, PLN probabilistic abstraction, and continuous PC modules for multimodal input. The result would be a system that, like GPT, can handle sequences and produce coherent outputs, but now grounded in a framework that actively minimizes prediction error, leverages causal reasoning to reduce hallucinations, and uses probabilistic logic to build deeper, more generalizable abstractions.

\subsection{Integrating Additional Memory and Reasoning Components}

If one is going to take the trouble to go this far and implement something transformer-like in an ActPC-Chem based framework, then one might as well go a bit further and upgrade the transformer-like architecture to include additional memory components that will allow it to serve as a more fully functional component of an integrated cognitive system (e.g perhaps a PRIMUS system with a vaguely LLM-like network as a subcomponent).

We elaborate here how an ActPC-Chem based transformer-like network, pursuing a "predict the next token" language modeling goal, could effectively incorporate explicit roles for long-term memory (LTM) and working memory (WM). In this approach suggested here, rewrite rules, adapted over time, shuttle information between these memory stores and shape the model's token predictions.

Extending the architecture from the prior section, we propose to replace standard transformer layers (with attention and feedforward sub-layers trained via backprop) with a network of rewrite rules whose probabilities and conditions are continuously updated by ActPC (discrete predictive coding). AIRIS introduces and refines causal rules ensuring logical and narrative coherence, while PLN creates probabilistic abstractions that generalize beyond observed patterns.  Here, however, we additionally propose to structure the system to have explicit long-term memory and working memory stores, and rewrite rules that move content between these memory stores, influencing the context used for next-token prediction.

\subsubsection{Memory Structures and Data Flow}

\paragraph{Long-Term Memory (LTM):}

The LTM component, as we envision it in a minimal "memory enhanced LLM like ActPC-Chem++" architecture,

\begin{itemize}
  \item Stores a large pool of rewrite rules (some highly abstract, others very specific), learned templates, and generalized patterns inferred over the course of training.
  \item Contains both AIRIS-derived causal rules and PLN-derived probabilistic abstractions that represent stable knowledge about language patterns and logical/narrative structures.
\end{itemize}

\paragraph{Working Memory (WM):}

Complementarily, in this sort of architecture the WM component

\begin{itemize}
  \item Holds a dynamically maintained subset of tokens, concepts, and intermediate inferences relevant to the current context (like the transformer's contextual embedding for the last N tokens, but now stored as a pattern of rewrite-applicable structures).
  \item Rewrite rules operating on WM determine which parts of the recently processed text or inferred concepts remain salient, which are combined, abstracted, or discarded.
  \item WM content can be viewed as a set of symbolic states reflecting the ongoing conversation: tokens, named entities, current topics, discourse structure, etc.
\end{itemize}

\paragraph{Rewrite Rules as Operators on Memory:}

To connect and utilize these memory stores,

\begin{itemize}
  \item There are rules for reading from LTM into WM (e.g., retrieving background knowledge relevant to the current conversation).
  \item There are rules for writing from WM back into LTM (e.g., if new stable patterns have emerged, store them as new long-term rules or increase the probability of existing rules).
  \item There are rules for transforming WM states to predict the next token. These are akin to the attention+feedforward steps in a transformer, but implemented as probabilistic rewrite steps that match current WM patterns to candidate next-token predictions.
\end{itemize}

\subsubsection{The Next-Token Prediction Cycle}

Given these additional memory structures, at each step of predicting the next token, we have the following processes:

\paragraph{ Input and WM State:}

\begin{itemize}
  \item The model receives the current partial text input: a sequence of tokens already generated or observed.
  \item The WM currently holds a symbolic representation of the last few tokens, extracted features, thematic concepts, and possibly inferred causal relationships relevant to the text (e.g., a character in a story, a topic under discussion).
\end{itemize}

\paragraph{Selecting Rewrite Rules for Next-Token Prediction (ActPC):}
\begin{itemize}
  \item A set of rewrite rules attempts to match the current WM state.
  \item Each rule may propose a candidate next token, or propose reading certain knowledge from LTM to refine the decision.
  \item ActPC manages the probability distribution over these rules, pushing more plausible rules (those that historically reduced predictive error) to be considered first.
\end{itemize}

\paragraph{Causal Reasoning (AIRIS) in WM:}
\begin{itemize}
  \item If certain transitions predicted by the rules conflict with the narrative logic observed so far, AIRIS steps in to refine or add causal conditions. For example, if the text implies a character cannot be in two places at once, but a rule tries to generate a token sequence violating this logic, AIRIS will introduce a condition preventing that rule's application unless it can be reconciled with a new explanation.
  \item Over time, AIRIS-derived rules ensure that the next-token predictions respect causal, logical, and narrative constraints rather than producing hallucinations that contradict previously stated facts.
\end{itemize}

\paragraph{Probabilistic Abstraction (PLN) for Generalization:}
\begin{itemize}
  \item PLN monitors patterns of tokens and the conditions in which certain next-token predictions were successful or not.
  \item Based on inductive reasoning, PLN may propose a new abstract rewrite rule: for example, if the model often sees that after certain discourse markers ("On the other hand, ...") the text structure follows a particular rhetorical pattern, PLN may create a generalized rule.
  \item This abstract rule then competes with or complements more specific rules, and through ActPC, the probability of applying this abstract rule increases as it proves useful.
\end{itemize}

\paragraph{Continuous PC Integration for Additional Modalities (Optional):}
\begin{itemize}
  \item If the textual input references an image or another modality, a continuous PC submodule processes that modality and updates WM with a symbolic feature representation of the image, audio, or sensor reading.
  \item Rewrite rules can incorporate these features from WM into next-token decisions, e.g., generating a token describing an image attribute.
\end{itemize}

\paragraph{Memory Interaction and Attention-Like Behavior:}
\begin{itemize}
  \item Instead of direct self-attention, rules can look up related patterns in WM and LTM:
  \item Retrieval rules from LTM to WM might say: "If the current topic is 'climate change' and we are discussing policy from 10 tokens ago, bring forth stored abstract rules about 'argumentation structure for policy discussion."'
  \item Thus, attention-like operations are carried out by rewriting: a rule matches the current WM pattern and says "pull in relevant knowledge from LTM," effectively focusing "attention" on certain background patterns.
  \item Similarly, if a certain piece of knowledge in LTM is frequently used given current WM patterns, ActPC will boost rules that retrieve it, making retrieval cheaper and more likely, mimicking how attention increases weights for relevant tokens in a transformer.
\end{itemize}

\paragraph{Updating the Rewrite Rules Themselves:}
\begin{itemize}
  \item After predicting the next token, the actual token arrives. ActPC computes prediction error: if the chosen rule's prediction was correct or close, it increases that rule's probability; if not, probability shifts away.
  \item If persistent errors occur in certain contexts, AIRIS introduces or refines rules with new conditions. PLN might generalize a different rule to cover these cases.
  \item Occasionally, stable new patterns discovered in WM can be written into LTM as a new permanent rule. For instance, if the model repeatedly finds that when a certain thematic pattern appears, the next tokens follow a known rhetorical sequence, it stores a new set of rules encoding this sequence as a stable pattern.
\end{itemize}

\paragraph{Long-Term Stability and Adaptation:}

\begin{itemize}
  \item Over many episodes of reading or generating text, the system builds a large library of rewrite rules in LTM, some focusing on low-level token transitions and others on high-level discourse and causal logic.
  \item AIRIS ensures that blatantly illogical transitions are pruned or restricted by adding conditional logic.
  \item PLN ensures that as the system encounters more texts, it can generalize and handle unseen constructions without brute-force memorization.
\end{itemize}

\subsubsection{Putting It All Together}

We have proposed a memory-enhanced transformer-like architectures intended for operation in an ActPC-Chem++ setting, involving:

\begin{itemize}
  \item A next-token prediction loop driven by rewrite rules adapting via ActPC.
  \item AIRIS continuously refining causal conditions to maintain logical/narrative consistency.
  \item PLN introducing probabilistic generalizations to handle new patterns.
  \item WM and LTM acting as dynamic contexts and knowledge bases, managed entirely by rewrite rules rather than static embeddings or learned matrices.
  \item Attention-like behavior emerges from rule-driven retrieval and relevance operations rather than dot-product computations.
\end{itemize}

\noindent This approach creates a transformer-like system but grounded in a fundamentally different learning paradigm, potentially reducing hallucinations and enhancing interpretability and logical consistency in language modeling -- and most critically, providing a clear way to connect the power of transformer-like dynamics to the other sorts of cognitive processing needed to make a human-like mind or an AGI/ASI with greater than human capabilities.

The potential benefits of this approach include:

\begin{itemize}
  \item Analogy to Transformer Layers: Each "layer" in a standard transformer can be replaced by a round of rewrite-rule applications on the WM state. Instead of Q-K-V attention, we have "retrieval rewrite rules" that pull relevant patterns from LTM into WM. Instead of MLP feedforward layers, we have "transformational rewrite rules" that interpret the current WM pattern and propose next tokens or intermediate conceptual tokens.
  \item Context and Coherence: Over multiple prediction steps, the WM acts like the transformer's context window, but rules can adjust what stays in WM longer or shorter than a fixed context length. Important narrative or logical elements can be retained longer via special rewrite rules that maintain them in WM or store them in LTM for quick retrieval when relevant again.
  \item Reducing Hallucinations: Hallucinations often stem from spurious correlations in large language models. Here, AIRIS identifies when a predicted token breaks causal/logical coherence and imposes conditions on rules to prevent recurrence of that illogical pattern. PLN can also abstract away from superficial patterns and focus on deeper logical or semantic regularities, guiding the model to more grounded predictions.
  \item Continuous Improvement: As reading and generation continues, new rules appear, old ones are pruned or refined, and probabilities adjust. The system's knowledge base evolves continuously, guided by ActPC error minimization, AIRIS causal logic, and PLN's higher-level abstractions.
\end{itemize}

\subsection{Approaches to Hierarchical Layering}

We now give a few more comments on how hierarchical layering might be worked out in such an architecture, combining the hierarchical nature of transformers with the multiple varieties of symbolic learning that can be integrated into the process in the ActPC-Chem setting.

In a transformer, we have multiple sequential layers, each containing attention and feedforward sub-layers. Similarly, we can organize our rewrite-rule architecture into multiple "layers." Each layer is a collection of rewrite rules that process the working memory (WM) state and produce a refined version of it. 

The lower layers might contain rules that handle simpler, more direct pattern completions, such as token-level co-occurrences or frequently seen local syntactic patterns. As we go up the layers, we encounter increasingly abstract rules. 

For example, we might say:

\begin{itemize}
  \item Lower Layer Rules: Deal with immediate local token predictions: "Given tokens ['The', 'cat', 'sat'], a likely next token is 'on'." These rules are relatively direct and rely mostly on local context.
  \item Mid-Layer Rules: Start to incorporate causal logic (AIRIS) or medium-range dependencies. They might reason: "Given we've established that the story involves a cat indoors, and previously we mentioned a sofa, the next likely token describing where the cat sat is 'on the sofa' rather than 'on the mat'."
  \item Higher Layer Rules (PLN and High-Level AIRIS): In upper layers, we apply rules that capture conceptual abstractions or narrative coherence. These might incorporate probabilistic logical inferences (PLN) and complex causal reasoning (AIRIS) to ensure that the generated text follows consistent, high-level patterns (e.g., character motivations, thematic consistency, or logical consistency across long sequences).
\end{itemize}

At each layer, ActPC ensures that the probability distributions over rules in that layer adapt based on prediction error and reward signals. Over time, the system "learns" which abstract patterns are beneficial and which aren't, just as a transformer learns higher-level representations in upper layers.

\subsubsection{Integrating With the Memory Structure:}

The dynamics of interaction between different layers in the network and the different memory stores may be interesting.   For instance, lower layers might primarily transform the current WM state based on local context, while higher layers might also pull information from LTM or refine conditions introduced by PLN and AIRIS. Thus, each layer can be viewed as a stage where the WM state is transformed, enriched, and made more coherent before moving on to the next layer.

\subsubsection{Attention Mechanism via Rule Selection}

In a standard transformer, attention selects parts of the input sequence to focus on using a dot-product-based mechanism that yields attention weights. In the rewrite-rule paradigm, instead of computing attention weights, each "attention-like" step involves a set of rules that look for patterns within the WM state (and possibly query LTM). For instance, a rule might say:

\begin{itemize}
  \item "If WM contains mention of a certain entity and a previously unresolved reference, retrieve related information from LTM."
\end{itemize}

The presence of a pattern that matches these conditions effectively acts like "attending" to that portion of the WM context. If the pattern is found and the rule is triggered, it brings relevant content or constraints to the forefront.

\paragraph{Probabilistic Selection as Attention Weights:} The probability of selecting a particular rule (guided by ActPC's prediction error minimization, along with other mechanisms such as ECAN) acts somewhat like an attention weight. Rules that consistently bring helpful context or correct predictions gain higher probability, making them more likely to be triggered in similar future contexts. Over time, the system learns a distribution of "which rules to attend to" for particular states.

\paragraph{Contextual Focus Emerges Dynamically:}  As different rules compete or cooperate, the system naturally "focuses" on the relevant segments of WM or LTM. This is akin to attention heads in a transformer: multiple sets of rules can operate in parallel, each focusing on different aspects of the context. Some rules might specialize in attending to named entities, others to temporal sequences, others to thematic consistency.

\subsubsection{Feedforward Sublayers as Rewrite Transformations}

A transformer's feedforward sub-layer takes the output of the attention sub-layer and transforms it through a series of nonlinear operations, producing a refined embedding.

To see how dynamics conceptually resembling the feedforward dynamics within transformers might emerge via ActPC-Chem rewrite rules:

\begin{itemize}
  \item In our architecture, the "feedforward" step can be represented by a set of rewrite rules that transform the current WM representation into a more predictive form. For example:
  \item After attention-like rules bring relevant context into WM, feedforward-like rewrite rules might reshape or combine tokens and concepts into a more structured pattern that leads directly to next-token prediction.
  \item These rules might insert intermediate symbolic concepts (similar to latent embeddings) that bridge low-level token patterns and higher-level narrative elements. For instance, "Given the context, transform the concept ['cat', 'location'] into a predictive pattern ['cat\_on\_sofa']"-a more specific concept that can lead to a suitable token prediction like "on the sofa".
  \item Compositionality and Nonlinearity: The feedforward step in a transformer is nonlinear. In the rewrite-rule paradigm, nonlinearity emerges from the conditional and probabilistic nature of rule selection. A rewrite rule doesn't just linearly transform embeddings; it might trigger a complex conditional pattern insertion or removal, effectively performing nonlinear transformations on the symbolic representation.
\end{itemize}

\paragraph{ActPC Refinement of Feedforward Rules: } As with attention-like steps, ActPC adjusts the probabilities and conditions of these feedforward-like rules over time. If certain transformations consistently reduce prediction error, they become more probable and stable. If they fail, the system modifies or replaces them, possibly guided by AIRIS (to insert logical conditions) or PLN (to generalize or specialize the transformations).

\subsubsection{Cohesive Integration across the Network}

Layer-by-Layer Processing, in this setting, could flow something like:

\begin{itemize}
  \item Start with the input tokens and some initial WM state.
  \item Layer 1 (Low-Level Rules):  Apply attention-like rewrite rules to find relevant local patterns. Then apply feedforwardlike rewrite rules to refine or combine these patterns, producing a WM state better prepared for prediction.

  \item Layer 2 (Intermediate-Level Rules):  Possibly incorporate AIRIS conditions to ensure causality or logic. Attention-like rules here might focus on medium-range dependencies (e.g., consistency with events mentioned above), and feedforward-like rules might add abstracted concepts derived by PLN.

\item  Layer N (High-Level Rules):  At the top layers, the rewrite rules might rely heavily on PLN-derived abstractions and AIRIS-induced causal patterns. The attention-like rules here focus on very high-level narrative or logical structure, pulling relevant background knowledge from LTM. Feedforward-like rules transform these abstractions into a final pattern leading to coherent, contextually appropriate next-token predictions.
\end{itemize}

\paragraph{Multiple Parallel Heads and Distributed Attention:} We can conceptualize multiple "heads" of attention as different sets of rewrite rules operating in parallel. Each set of rules competes or cooperates to transform the WM, capturing different facets of the context (e.g., semantic similarity, syntactic role, narrative continuity).

\paragraph{End-to-End ActPC Updates:} At the end of the next-token prediction cycle, the observed token is compared with the predicted distribution. ActPC's error-driven update affects:

\begin{itemize}
  \item Which attention-like rules get more or less probable,
  \item Which feedforward-like transformations are favored,
  \item How conditions introduced by AIRIS and abstractions from PLN affect future rule application.
\end{itemize}

Over time, in this way, the entire stack of rewrite-rule layers evolves to produce increasingly accurate and contextually rich next-token predictions, mimicking how a transformer's layers learn richer representations but now driven by a continuous cycle of local pattern matching, probabilistic selection, and error-driven adjustment rather than backprop gradients.

\subsubsection{Hopeful Summary of the Architecture}
By structuring the rewrite rules into stacked layers (mirroring transformer layers), using pattern matching and rule selection as a form of attention, and employing conditional rewriting operations as a feedforward step, we replicate the functional components of transformer layers in a probabilistically-driven, non-backprop paradigm. AIRIS ensures causal/logical coherence, PLN helps form useful abstractions, and ActPC continuously tunes the probability distribution over rules for all these processes.

\subsection{Rough Architecture Visualizations}

\subsubsection{Language Processing Architecture Sketch} 

Integrating many of these ideas, below is an ASCII-style architecture diagram illustrating the transformer-like NLP system described. It's a high-level conceptual diagram, capturing the main components (Layers, WM, LTM, Rule Sets) and their interactions.  While crude, it does provide a structured overview of  some of the key the modules and data flows.

\includegraphics[width=10cm]{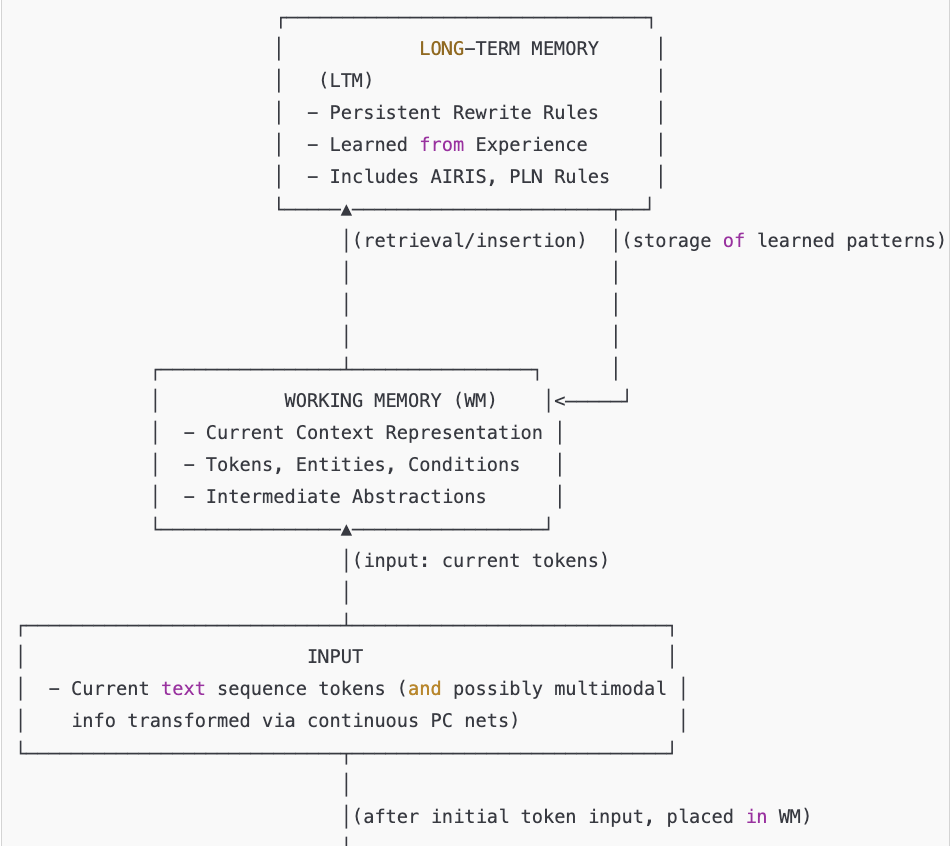} \\
\includegraphics[width=10cm]{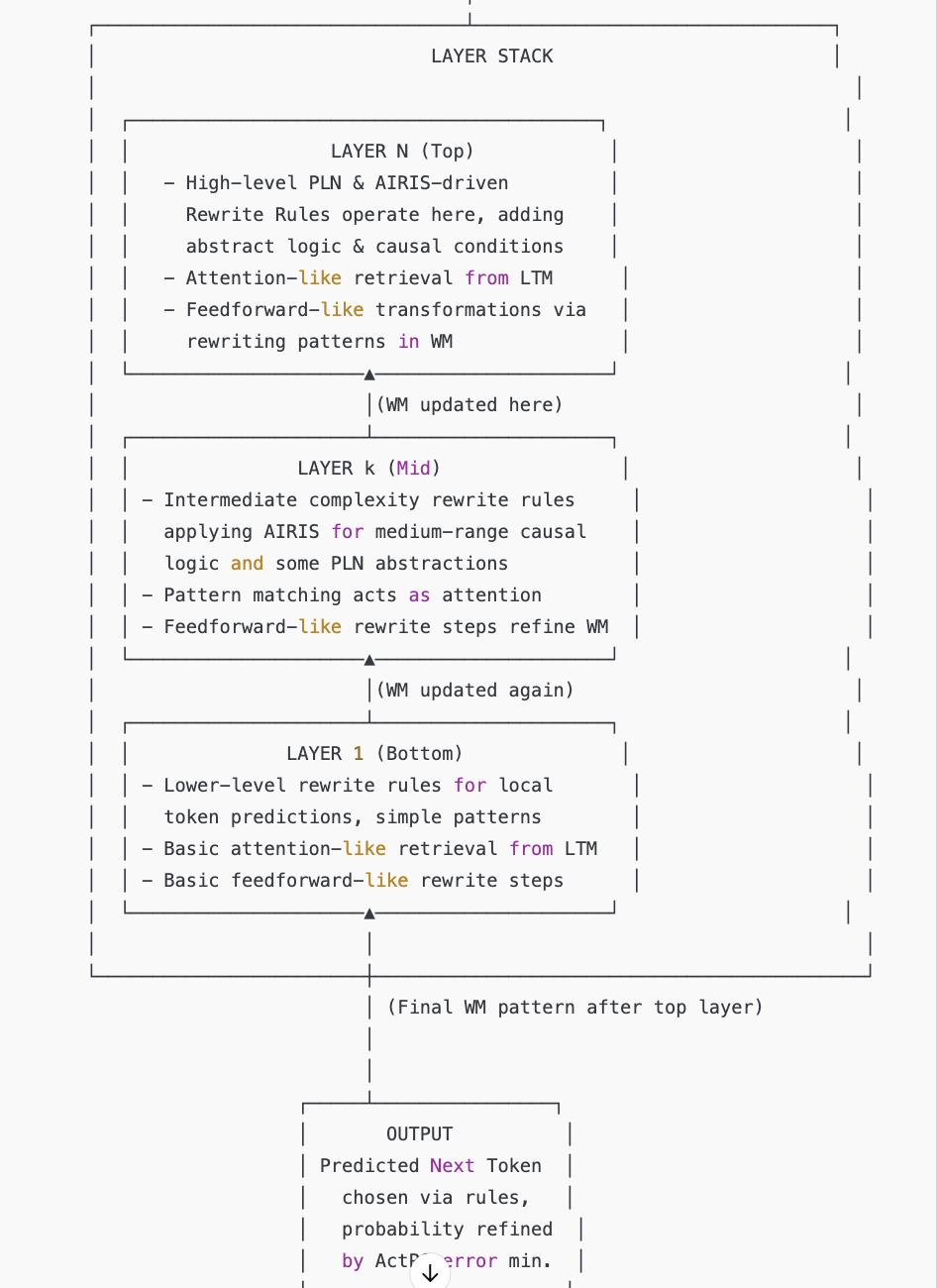}

Key Concepts:

\begin{itemize}
  \item Each layer consists of sets of rewrite rules that:
  \item "Attend" to parts of WM and possibly retrieve relevant patterns from LTM (Attention-like).
  \item Apply transformations to WM states (Feedforward-like).
  \item Are updated probabilistically by ActPC to minimize prediction error.
  \item Higher layers incorporate AIRIS for causal logic and PLN for abstract probabilistic reasoning.
  \item The WM (Working Memory) is repeatedly updated as we pass through layers, gradually refining the representation until the system produces a next-token prediction.
  \item LTM (Long-Term Memory) provides a store of learned rewrite rules, causal conditions (AIRIS), and abstract patterns (PLN). Layers may query LTM to bring additional context into WM.
  \item ActPC continuously adjusts the probabilities of applying certain rules based on prediction errors, ensuring the system evolves toward more accurate and coherent predictions.
\end{itemize}

This diagram shows the data flow from input tokens into WM, through multiple layers of rewrite-rule transformations (replacing standard transformer layers), culminating in a next-token output. The integration of AIRIS and PLN occurs mostly in the higher layers and in the structure of rules stored in LTM.

\subsubsection{Talking Robot Bug Architecture Sketch} 

Extending and complementing the above diagram and in a similar spirit, below is an ASCII-based conceptual diagram integrating all the discussed components into one "conversing robot bug" software architecture. 

This diagram is high-level, focusing on modules and data flow rather than code-level details. It shows how traditional predictive coding (PC) networks handle perception and actuation, while discrete ActPC, AIRIS, PLN, and memory structures (WM, LTM) support symbolic reasoning, language processing, and causal inference. The layered transformer-like architecture for language is also included, along with multimodal integration.

\includegraphics[width=10cm]{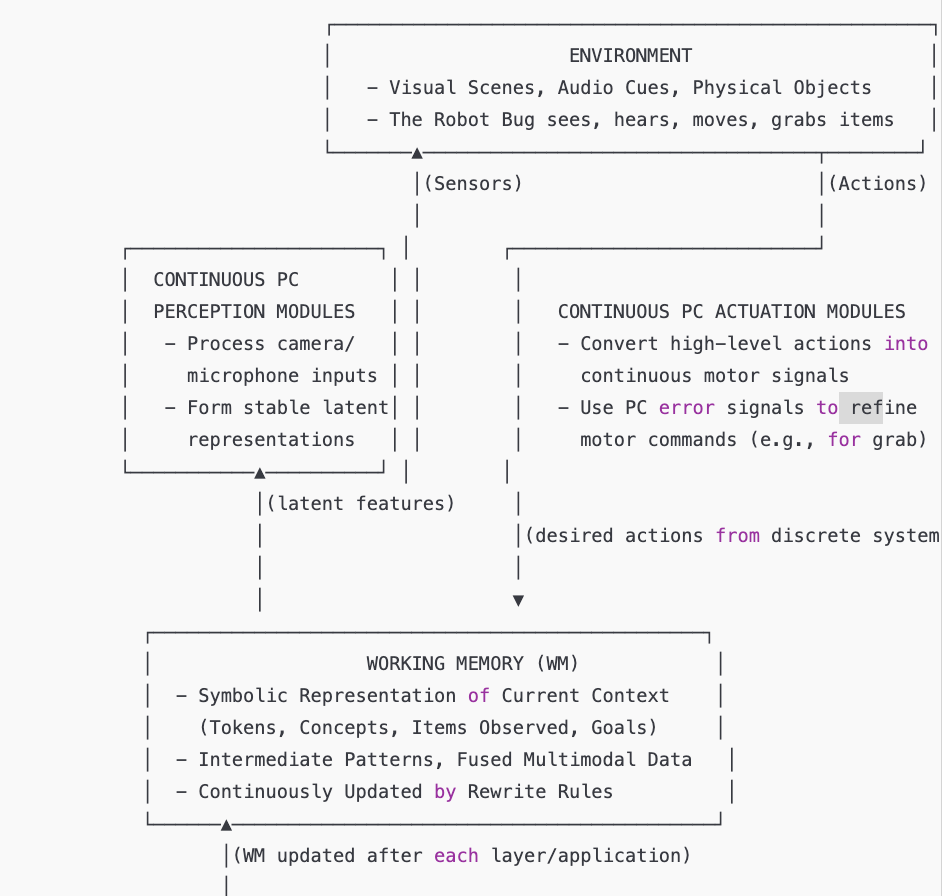} \\
\includegraphics[width=10cm]{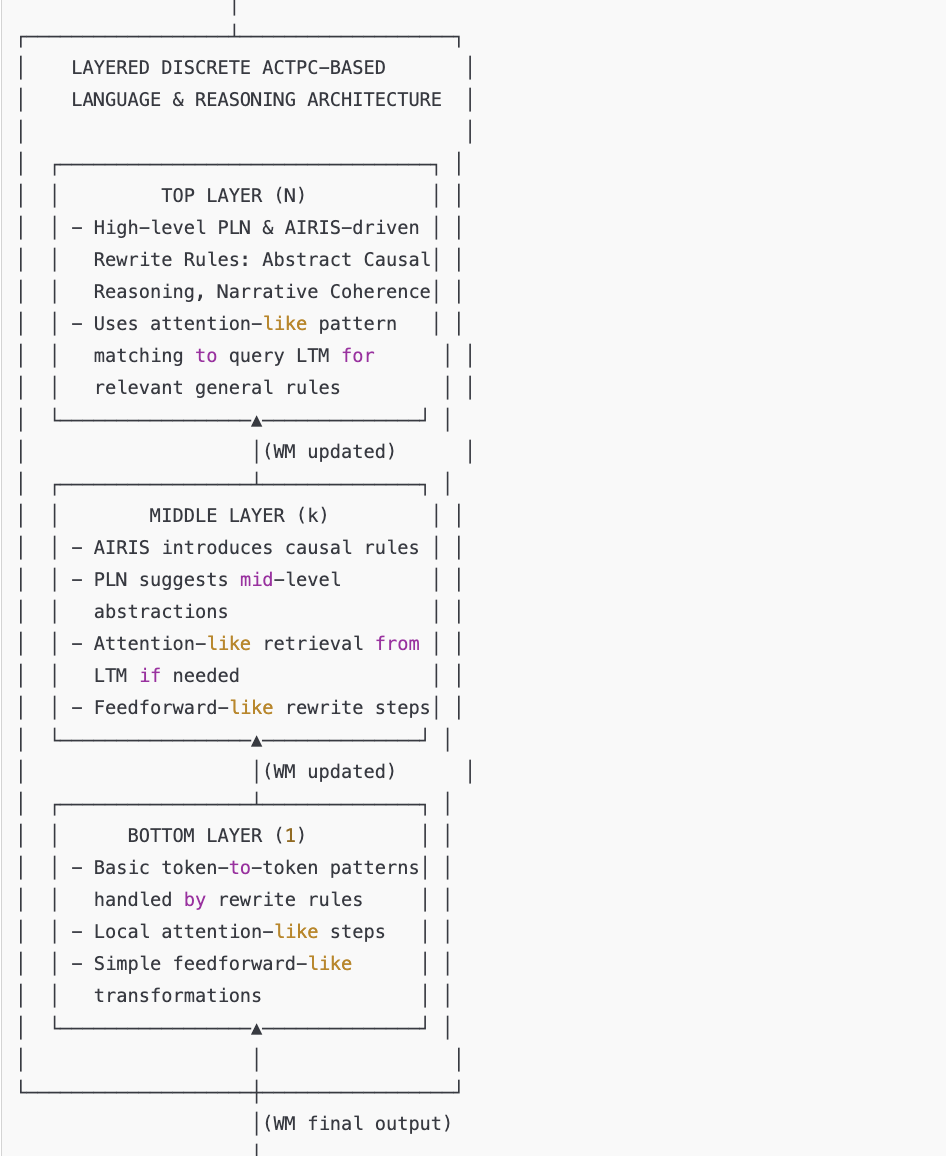} \\
\includegraphics[width=10cm]{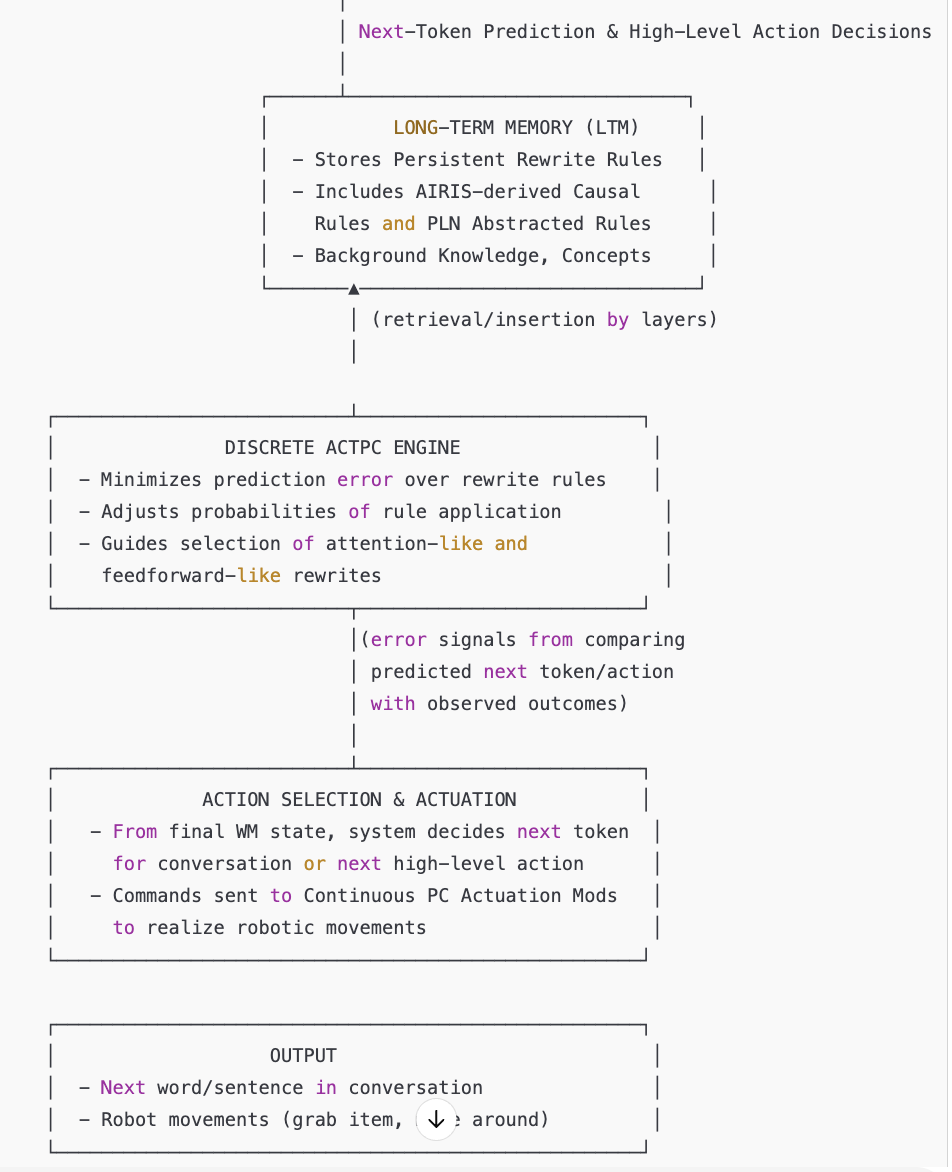} \\

Notes:

\begin{itemize}
  \item Continuous PC modules handle raw sensory input (vision, audio) and motor signals, producing stable latent features/receiving action commands.
  \item WM holds the current linguistic/contextual and conceptual state of the conversation and scenario.
  \item Layers of discrete ActPC-driven rewrite rules perform functions analogous to transformer attention and feedforward, enriched by AIRIS (causal rules) and PLN (abstract inference).
  \item LTM stores long-term learned patterns, rules, and abstractions, which areretrieved into WM as needed.
\item The entire system's choice of rules is driven by ActPC's error minimization,
ensuring continuous adaptation and improvement.
\item The robot bug can converse, referencing its perceptions and actions; the
integration ensures logically coherent and contextually rich responses.
\end{itemize}

This diagram shows the integrated system: sensor and motor loops at the bottom (via continuous PC), a discrete rewrite-rule-based "transformer-like" stack in the middle (with AIRIS, PLN, and ActPC guiding rule probabilities), and memory structures (WM and LTM) enabling context, abstraction, and causal logic. The result is a coherent multi-level architecture where language production, perception, action, and logical inference all mesh together.

\subsection{ Integrated Predictive Error Handling Across Language, Perception, Action and Cognition}

We wrap up this section on transformer-like ActPC-Chem architectures with some speculative analysis of integrated predictive error handling across the whole network governing a hypothetical "talking robot bug" operated using ActPC based components throughout, including transformer-like rewrite rule networks, rewrite rule networks for cognition and planning, and traditional ActPC neural networks for perception and action.

In this integrated ActPC-Chem approach, the entire cognitive stack of the robot bug---encompassing perception, action, and transformer-like language generation---is grounded in predictive coding principles and implemented through a network of evolving rewrite rules. Each subsystem (perception, action, language) maintains internal probability distributions over sets of rules or continuous feature distributions, all of which are adapted online, in real time, via local prediction error minimization. This unified predictive-coding-based substrate ensures that errors detected in one subsystem can be coherently propagated and addressed in conjunction with errors arising in other subsystems.

\subsubsection{Integrated Error Management}

Consider for instance the scenario where the robot bug verbally requests a specific piece of ``red'' food from a collaborator, gesturing with its grabber. The collaborator returns a different piece of food than intended. Several sources of prediction error arise:

\paragraph{Perceptual Prediction Error:}. The bug's vision system, implemented as a continuous predictive coding (PC) subnetwork, had previously assigned a high probability to categorizing the item as ``red.'' Given the outcome, it becomes apparent that the color perception was too coarse. The correct categorization might need to be ``light red'' or ``pinkish-red,'' reflecting a finer-grained perceptual classification. To correct this, the continuous PC network that encodes visual features $z_{\text{vis}}$ will adjust its latent distributions:

\[
e_{\text{perception}} = D_{KL}\bigl(q(\text{features}\mid\text{light-red}) \,\|\, p(\text{features}\mid\text{red})\bigr)
\]

Here, $D_{KL}$ denotes the Kullback--Leibler divergence (or a similar measure) quantifying the mismatch between the observed spectral properties of the food item and the previously hypothesized ``red'' category. The PC inference process will then nudge the internal representations and rule probabilities associated with color categories toward a refined color taxonomy.

\paragraph{Action Prediction Error:}. The bug's motor system, another continuous ActPC network, made predictions about how a certain arm gesture would appear to the collaborator. If the collaborator misidentified the referenced food because the gesture was ambiguous from their vantage point, the action rules or continuous motor control distributions need adjusting. The transformation that maps the intended pointing direction $a$ to the observed collaborator response yields a prediction error:

\[
e_{\text{action}} = D_{KL}\bigl(q(\text{collaborator\_interpretation}\mid a) \,\|\, p(\text{collaborator\_interpretation}\mid a)\bigr)
\]

Minimizing this error involves reconfiguring the rewrite rules or continuous control distributions to incorporate perspective-taking: perhaps the bug needs to shift its pointing angle or add an additional clarifying movement. Since all action selection is guided by predictive coding, the system can experimentally vary action rules and select those that reduce future action-related errors.

\paragraph{Language Prediction Error:}. The language subsystem, implemented as a discrete ActPC-chem transformer-like network, introduced a rule that produced the phrase ``that red food over there.'' Given the final outcome, it is clear that this phrasing was not sufficiently discriminative. The language model thus encounters an error:

\[
e_{\text{language}} = D_{KL}\bigl(q(\text{outcome}\mid\text{``that red food over there''}) \,\|\, p(\text{outcome}\mid\text{``that red food over there''})\bigr)
\]

Here, the ``outcome'' includes the collaborator's incorrect item selection. The discrete rewrite rules that govern reference generation and descriptive phrasing must now be updated. More precise descriptors---e.g., referencing size, shape, or spatial location more explicitly---might reduce this error.

\paragraph{Coordinated Correction Across Subsystems:}. Because each layer---perception, action, language---uses a form of predictive coding and handles distributions over rewrite rules or latent feature spaces, prediction errors feed back into local rule probability shifts. The key is that these subsystems are not isolated. The robot's cognitive architecture relies on the same underlying ActPC-chem principles everywhere, so errors from the language layer can influence which perceptual distinctions to emphasize or which action rules to update. Conversely, improved perceptual categorization or more perspective-aware gesturing can feed back to the language layer, prompting it to refine the lexical and syntactic rewrite rules used for object references.

\paragraph{Role of PLN and Analogical Reasoning:}. Probabilistic Logic Networks (PLN) provide a mechanism to introduce analogical and abductive reasoning into this process. Suppose the language subsystem struggles to choose the right descriptive phrase. PLN can search long-term memory (LTM) for past situations in which similar requests were made successfully. For example, if previously saying ``the small red apple on the left side of the table'' led to correct item selection, PLN can propose analogous patterns for the current scenario. PLN might infer that, since ``adding spatial qualifiers'' worked before, similar expansions (e.g., ``the lighter red piece of fruit near the corner'') should be tried now.

Formally, PLN might compute a probability that a certain descriptive pattern will lead to correct interpretation based on similarity to prior successful patterns:

\[
p(\text{success}\mid\text{new\_descriptor}) = f(\text{analogical\_similarity},\text{past\_successes})
\]

By considering analogies and known successes, PLN generates candidate variations of language rules. These candidate variations are then tested by ActPC: each candidate rule is assigned an initial probability, and the system experimentally applies these rules in future interactions. Outcomes feed back as prediction error signals that raise or lower the probabilities of these PLN-inspired variants.

\subsubsection{Integrated Adaptive Cycle}

The overall adaptive process involved here, identifying and responding to predictive errors and triggering cognitive processes in various modules and at various levels as a result, looks roughly like:

\begin{itemize}
\item  \emph{Error Identification:} The collaborator's incorrect response triggers a top-down re-evaluation: The system identifies that something about color naming, gesturing, or phrasing was off.

\item \emph{Local Adjustments:} Each subsystem (visual features, action trajectories, linguistic descriptors) tries local rewrites or continuous distribution updates to better align predictions with reality. These updates occur continuously and locally via ActPC, without global backpropagation.

\item \emph{PLN Proposals:} PLN reviews the memory of past successful communications and proposes new linguistic (or even action-related) patterns analogous to previous successes. These proposals appear as additional rewrite rules or modifications to existing ones.

\item \emph{Experimental Variation and Probability Shifting:} The system attempts these new PLN-suggested patterns in subsequent requests. If prediction errors diminish, these variants gain higher probability and become stable policies. If not, they are discarded or refined further.
\end{itemize}

Because the entire architecture relies on probability distributions and prediction error minimization at multiple levels---continuous PC for perception and actuation, discrete ActPC-chem for cognition and language, and PLN for guided analogical variations---the correction of errors can be coordinated across all layers. Instead of patching one component at a time, the system discovers correlated improvements that harmonize perception, action, and language. This holistic error-correction cycle endows the robot bug with robust adaptability, able to refine its behavior incrementally and coherently as it interacts with the world and other agents in real-time.

\section{Conclusion}

In this paper, we have presented a speculative yet moderately well fleshed out and conceptually very rich approach to creating a flexible, hybrid AI architecture called ActPC-Chem, which we envision as a potential "cognitive kernel" for future AGI systems, fitting in especially closely with variations of the PRIMUS cognitive architecture.

By grounding both data and models in a metagraph of rewrite rules, and using Active Predictive Coding (ActPC) principles as the primary adaptation mechanism, we have outlined a path toward integrating continuous sensorimotor control with discrete symbolic and causal reasoning.   Implementation-wise, this sort of large-scale rewrite rule based system fits very naturally with the MeTTa language and other components of the OpenCog Hyperon system.  Conceptually, we may say the approach is inherently open-ended and autopoietic --capable of continuously inventing and refining new patterns of thought and behavior -- while still capable of being guided by goal-driven reinforcement signals and structured by logical constraints.

We have explored how this approach could be used to control virtual or robotic experientially learning agents -- using animated or robotic bugs as examples.  We have argued that this setup naturally facilitates the seamless integration of symbolic reasoning (e.g., via AIRIS for causal inference and PLN for probabilistic logical abstraction) on top of a substrate that already handles perception and action through ActPC-based continuous PC networks. 

We have also explored how a transformer-like network can be adapted into this framework, replacing backpropagation-based weight updates with locally applied rewrite rules and probability shifts driven by predictive coding error signals. 

Each of these explorations fits very naturally with the potential use of ActPC-Chem within the PRIMUS architecture implemented on the Hyperon substrate, though they are also compatible with other choices of cognitive architecture and implementation framework.

The overall idea is a holistic cognitive architecture where perception, action, and language can be co-adapted in real time, guided by predictive error minimization and informed by analogical reasoning and causal logic. Such a system promises to yield robust, context-sensitive, and internally coherent behavior without resorting to massive, offline training cycles, and without requiring unrealistic amounts of computational resources.

Looking forward, there are numerous key directions for future work, including:

\begin{itemize}
\item {\bf Theoretical Refinement:} While the principles outlined here are plausible and conceptually appealing, deeper mathematical and algorithmic formalisms are needed. E.g. 

\begin{itemize}
\item further formalizing the discrete natural gradient updates and optimal transport-based geometry
\item formalizing the interrelation between discrete and continuous ActPC networks
\item formalizing and further analyzing the role of ECAN-driven rule selection in ActPC-Chem
\item formalizing and further analyzing the probabilistic causal and inferential modeling done in PLN and AIRIS with discrete-ActPC error measurement and correction
\item understanding which subsets of GSLT are most promising for experimentation with increasingly more complex base-level rewrite rules for ActPC
\item attempting to rigorously analyzing convergence properties and computational complexity of the overall system under various assumptions
\end{itemize}

\noindent  will all be valuable to ensuring that ActPC-Chem networks can scale to real-world complexity.

\item  {\bf Practical Implementation:} Implementing ActPC-Chem in actual code and integrating it into the OpenCog Hyperon platform will be a significant step toward turning these ideas into a workable system. Incremental prototypes can be tested on simpler tasks within virtual environments, allowing stepwise validation of each component -- continuous PC perception/action, discrete rewrite-rule adaptation, AIRIS causal logic, PLN-driven analogical inference, and transformer-like token prediction.

\item  {\bf Gradual Expansion Toward PRIMUS-Based AGI:} As the kernel matures, more elements of the PRIMUS cognitive architecture can be layered on top. Over time, one can incorporate richer commonsense knowledge bases, structured world models, and increasingly sophisticated forms of memory and reasoning. By starting small -- perhaps with virtual-world scenarios -- developers can gradually add complexity until a framework capable of yielding a robust, human-level AGI emerges.

\item  {\bf Experimental Evaluation in Virtual and Physical Realms:} The Sophiaverse virtual world, and particularly its Neoterics sub-world, provides an ideal environment for early experimentation. Agents can be trained and tested in these controlled yet richly interactive virtual spaces, allowing for rapid iteration and refinement of the ActPC-Chem principles. Subsequently, translation into real-world robotics -- such as the Mind Children humanoid robots -- would test the approach in noisy, dynamic, and physically embodied settings.
\end{itemize}

If successful, this line of research could play a major role in the inevitably emerging paradigm shift away from monolithic backprop-trained networks toward more evolving, self-organizing cognitive architectures. Such architectures, if built with ActPC-Chem as a core ingredient, would unify perception, action, language, logic, and abstraction within a single predictive-coding-oriented framework, opening the door in this way to genuinely flexible, adaptable, and eventually AGI and ASI-level intelligence.

\section*{Acknowledgement} As well as my Hyperon/PRIMUS collaborators and Alex Ororbia and my family and the multiverse in general, I'd like to thank Daniel McDonald for putting into my head the idea of exploring discrete variations of Orobia's Predictive Coding based learning.

\bibliographystyle{alpha}
\bibliography{actpc}

\end{document}